\PassOptionsToPackage{table,xcdraw,rgb}{xcolor}
\documentclass[table]{article}
\PassOptionsToPackage{numbers, sort}{natbib}
\usepackage{iclr2025_conference,times}

\usepackage[utf8]{inputenc} 
\usepackage[T1]{fontenc}    
\usepackage{hyperref}
\usepackage{etoolbox}
\usepackage{url}            
\usepackage{booktabs}       
\usepackage{array}
\usepackage{amsfonts}       
\usepackage{nicefrac}       
\usepackage{microtype}      
\usepackage{graphicx}
\usepackage{amsmath}
\usepackage{algorithm}
\usepackage{algpseudocode}
\usepackage{multicol}  
\usepackage{breakcites}
\usepackage{makecell}
\usepackage{wrapfig}
\usepackage{caption}
\usepackage[font=footnotesize,labelformat=simple]{subcaption}
\usepackage{subcaption}

\makeatletter
\newcommand{\thickhline}{%
    \noalign {\ifnum 0=`}\fi \hrule height 1pt
    \futurelet \reserved@a \@xhline
}

\definecolor{lightgreen}{rgb}{0.25098039, 0.63921569, 0.34901961}
\definecolor{lightblue}{rgb}{0.6784313725, 0.803921569, 0.8196078431}
\definecolor{lightgray}{rgb}{0.90196079, 0.90196079, 0.90196079}
\algrenewcommand{\algorithmiccomment}[1]{\hfill\textcolor{lightgreen}{\(\triangleright\) #1}}

\makeatletter
\patchcmd{\@citex}{\@citea}{\def\@citea{\ifhmode\unskip\fi\space}}{}{}
\makeatother

\hypersetup{
    breaklinks=true     
}

\usepackage{natbib}
\usepackage{times}
\usepackage{epsfig}
\usepackage{graphicx}
\usepackage{amsmath}
\usepackage{amssymb}
\usepackage{booktabs}
\usepackage{multirow}
\usepackage{comment}
\usepackage{xcolor}
\usepackage{comment}
\usepackage{longtable}
\usepackage[accsupp]{axessibility} 

\usepackage{xspace}

\newcommand{\dd}[1]{#1}
\title{GALA: Geometry-Aware Local Adaptive Grids for Detailed 3D Generation}

\author{Dingdong Yang$^1$ \quad
Yizhi Wang$^1$ \quad
Konrad Schindler$^2$ \quad
Ali Mahdavi-Amiri$^1$ \quad
Hao Zhang$^1$ 
\\ \vspace{2px}
$^1$Simon Fraser University. \quad
$^2$ETH Zurich. \\
}

\usepackage{xspace}
\newcommand{\acro}{GALA\xspace}
\iclrfinalcopy

\begin{document}

\maketitle

\begin{figure}[htbp]
    \centering
    \includegraphics[width=\textwidth]{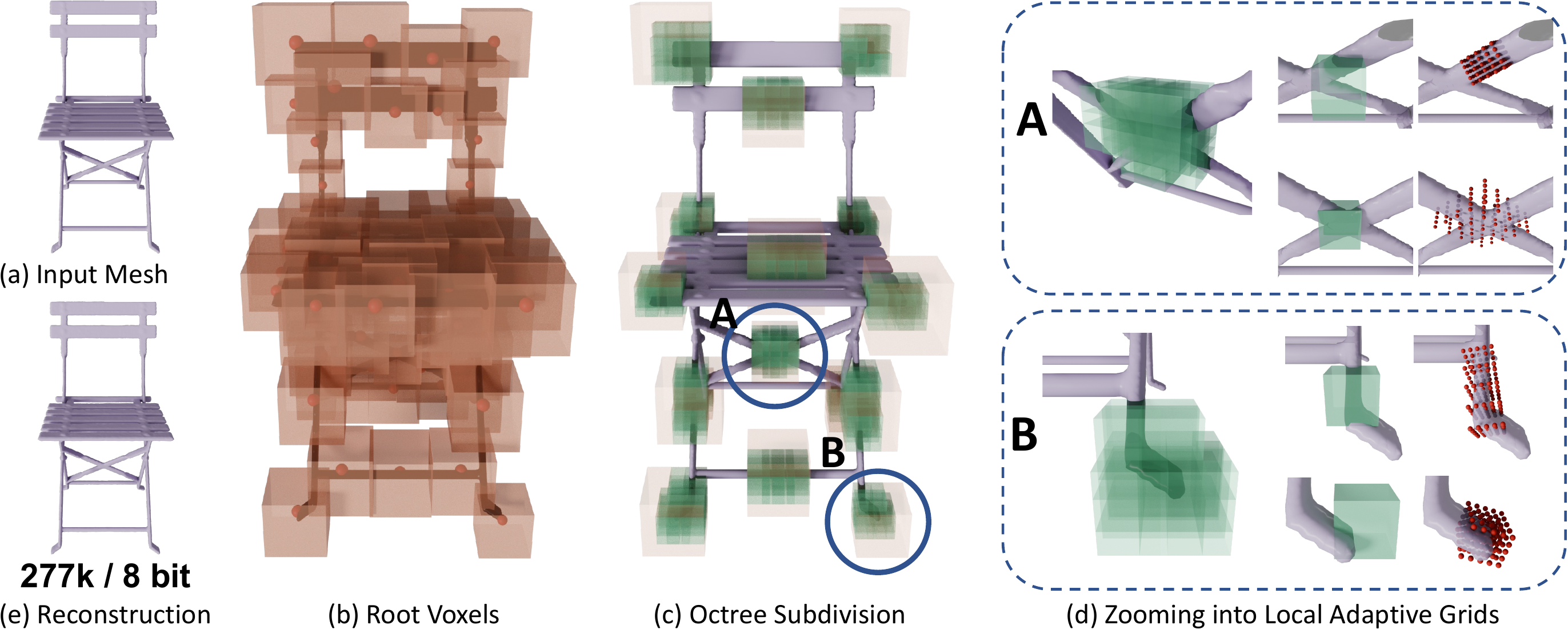}
    \caption{Given a watertight mesh (a), our representation, GALA, for {\em geometry-aware local adaptive grids\/}, distributes a set of root node voxels (\textcolor[rgb]{0.92157, 0.36863, 0.20392}{coral}) to cover the mesh {\em surfaces\/}. An octree subdivision is applied to each root, with a subset shown in (c). In each {\em non-empty\/} octree leaf node (\textcolor[rgb]{0.203922, 0.921568, 0.50196}{green}), a local grid (\textcolor[rgb]{1 0 0}{red dots}) is oriented and anisotropically scaled to adapt to and tightly bound the local surface geometries. Only 277K parameters with 8-bit quantization   yields an accurate representation (e).}
    \label{fig:most_front_illusration}
\end{figure}

\begin{abstract}
We propose GALA, a novel representation of 3D shapes that (i) excels at capturing and reproducing complex geometry and surface details, (ii) is computationally efficient, and (iii) lends itself to 3D generative modelling with modern, diffusion-based schemes. The key idea of GALA is to exploit both the global sparsity of surfaces within a 3D volume and their local surface properties. \emph{Sparsity} is promoted by covering only the 3D object boundaries, not empty space, with an ensemble of \dd{tree root voxels}. Each \dd{voxel} contains an octree to further limit storage and compute to regions that contain surfaces. \emph{Adaptivity} is achieved by fitting one local and geometry-aware coordinate frame in each \dd{non-empty leaf node}. Adjusting the orientation of the local grid, as well as the anisotropic scales of its axes, to the local surface shape greatly increases the amount of  detail that can be stored in a given amount of memory, which in turn allows for quantization without loss of quality. With our optimized C++/CUDA implementation, GALA can be fitted to an object in less than 10 seconds. Moreover, the representation can efficiently be flattened and manipulated with transformer networks. We provide a cascaded generation pipeline capable of generating 3D shapes with great geometric detail.
\end{abstract}

\begin{figure}[t!]
    \vspace{-4pt}
    \centering
    \includegraphics[width=\textwidth]{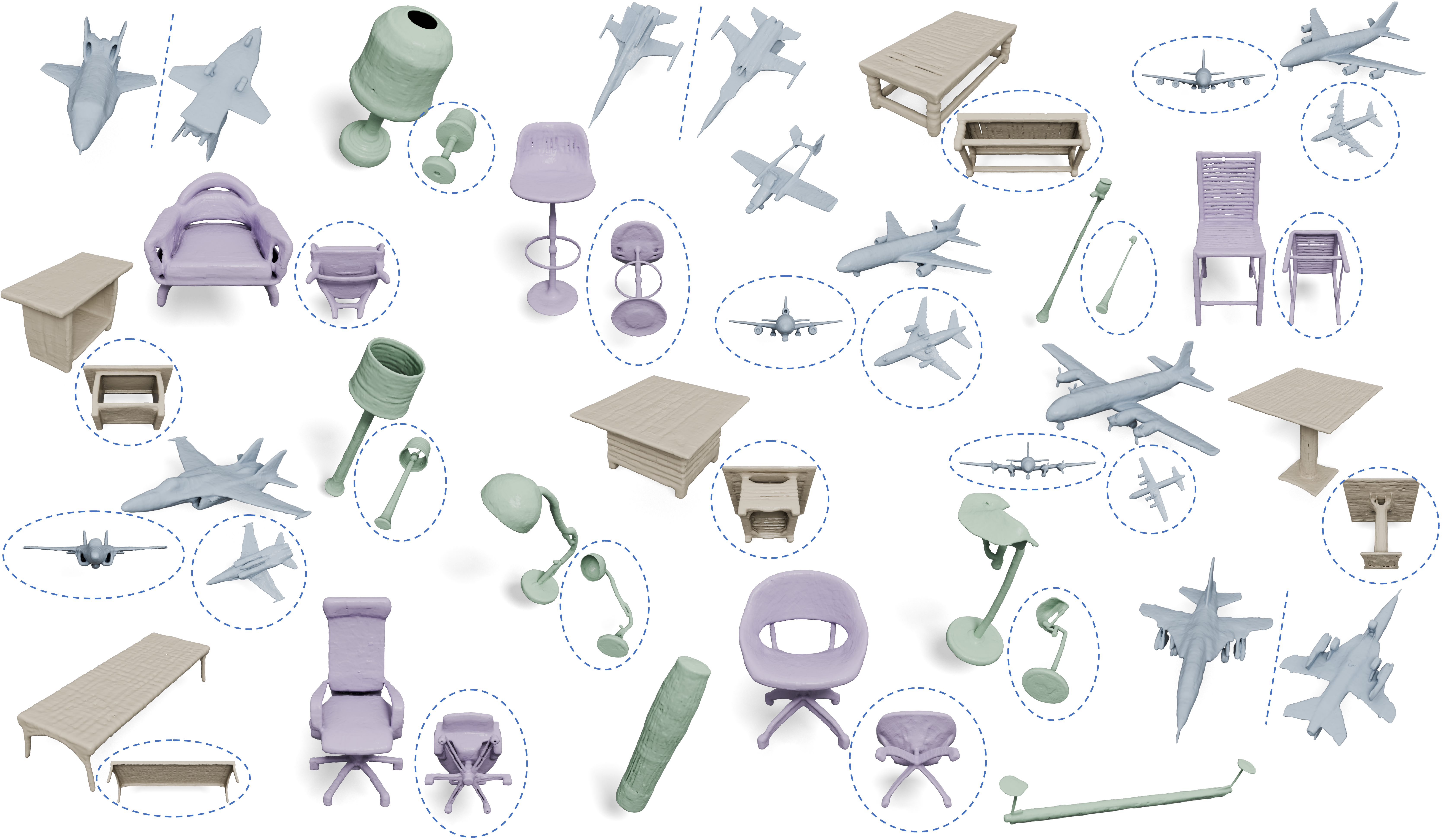}
    \caption{GALA enables diverse and detailed conditional 3D shape generation, including \textbf{\textcolor[rgb]{0.61568 0.713725 0.8313725}{Airplanes}}, \textbf{\textcolor[rgb]{0.615686274509804 0.8313725490196079 0.6549019607843137}{Lamps}}, \textbf{\textcolor[rgb]{0.8313725 0.7490196 0.6156872}{Tables}} and \textbf{\textcolor[rgb]{0.7254590 0.6156827 0.8313725}{Chairs}}. Best viewed on screen with high magnification.}
    \label{fig:teaser}
\end{figure}

\section{Introduction}
\label{sec:intro}

The generation of high-quality 3D assets remains a costly task in terms of processing time and resources involving human efforts or machine compute. 
%
%
A continuing research pursuit is to develop efficient 3D representations to streamline the traditionally expensive 3D creation workflow\linebreak[4] ~\cite{gupta20233dgen,jun2023shap,li2023diffusion,nichol2022point,shue20233d}, with the ultimate goal of speeding up the generative process without compromising the diversity and quality of the generated assets as reflected by their structural and geometric details.

To date, a variety of 3D representations including voxel~\cite{li2023diffusion} and tetrahedral grids~\cite{kalischek2022tetrahedral}, meshes~\cite{liu2023meshfusion}, point clouds~\cite{luo2021diffusion,zhou20213d}, neural fields~\cite{jun2023shap}, and triplanes~\cite{chan2022efficient,shue20233d,gupta20233dgen}, have been employed for 3D generation, often leveraging powerful diffusion-based generators.
%
%
However, as argued recently by~\cite{yariv2023mosaic}, none of these representations fulfill all of the three critical criteria for high-quality and scalable 3D generation:
parameter efficiency for detail representation, preprocess efficiency for handling large datasets and
mesh conversion, as well as simplicity of a 
tensorial form to suit modern-day neural processing.
In response to this, they proposed Mosaic-SDF which
encodes a signed distance function using a ``mosaic'' of aligned volumetric grids distributed over the shape surface, with different centers and \dd{different} (still \dd{isotropic}) scales.

In this paper, we accentuate that the key to 3D representation efficiency when using such an ensemble-of-tiles approach is {\em adaptivity\/} of the local coordinate grids to the geometry of the presented shapes. To this end, we propose to represent a 3D shape using a set of {\em geometry-aware\/} and {\em local adaptive\/} grids. Our novel representation, coined \acro, pushes the limit of adaptivity on three fronts:
\begin{itemize}
\vspace{-3pt}
\item To exploit the global sparsity of typical object surfaces within a 3D volume, we place a {\em forest of octrees}, rather than uniform subdivisions, at the object boundaries, where a local grid is associated with each {\em non-empty\/} octree leaf node to store SDF values.
\vspace{-2pt}
\item Our local grids are not globally aligned. Instead, each grid may be differently oriented based on local PCA analysis of surface normals within the corresponding octree subdivision.
\vspace{-2pt}
\item Last but not least, we {\em anisotropically} scale each grid to better capture geometric details.
\vspace{-3pt}
\end{itemize}



Our \acro representation can be implemented to yield sets of vectors that can be easily processed by transformer-based neural networks~\cite{vaswani2017attention}, while the octree forest provides a hierarchical 3D representation with limited depths. 
In ~\autoref{tab:compare_property}, we compare \acro with several state-of-the-art representative methods over several useful criteria.

The contributions of our work are threefold:

\begin{enumerate}
\item Our geometry-aware, locally adaptive, and anisotropic grids enable more efficient (\autoref{tab:compare_property}) and accurate sampling of shape structures, capturing geometric details, such as thin strings and plates, which may be missed by both regular grid-based methods~\cite{hui2022neural,li2023diffusion} and Mosaic-SDF, whose grid tiles are neither shape- nor detail-aware.
\item We have developed an efficient implementation in pure CUDA and libtorch\footnote{https://pytorch.org/cppdocs/} for the entire \acro extraction process. This process costs about 10 seconds for each watertight input mesh on Nvidia A100 GPU, which is among the fastest ones of popular fitting methods, enabling fast data processing for large datasets.
\item The hierarchical structure of the octree forest in \acro facilitates the development of a practical cascaded 3D generation scheme with improved efficiency and quality.
\end{enumerate}

\begin{table}[ht]
\centering
\resizebox{\textwidth}{!}{
\begin{tabular}{@{}lccccc@{}} 
\thickhline
\addlinespace[4pt] 
Methods & Description of representation & \makecell{Parameter \\ Count $\downarrow$} & \makecell{Fitting \\ Time $\downarrow$} & \makecell{Representation \\ Precision} & \makecell{Generation \\ Network} \\ 
\midrule
S2VS~\cite{zhang20233dshape2vecset} & Autoencoded vector set & $\leq$ 262k & - & Meidum/Low & Transformer \\
MeshDiffusion~\cite{liu2023meshdiffusion} & Deformable tetrahedral grid & $\geq$ 1.05M & 20m-30m & High/Medium & 3D UNet \\
Neural Wavelet~\cite{hui2022neural} & Two levels of wavelet coefficients & 536k & 1s & Medium & 3D UNet \\
Mosaic-SDF~\cite{yariv2023mosaic} & Grids on boundary & 355k & 2m & High & Transformer \\
\acro (ours) & Trees with adaptive grids on boundary & 277k & 10s  & High & Transformer \\
\addlinespace[4pt] 
\thickhline
\end{tabular}%
}
\caption{Properties of different representations for 3D generation. \dd{The representation precision is defined as the reconstruction errors with respect to input meshes. See~\autoref{sec:recon} for more.
 }}
\label{tab:compare_property}
\end{table}

\section{Related Work}

\subsection{3D Representation for Generation}

We will first review the related work on 3D representation for generation, focusing on various methods characterized by shared features.

\textbf{Grid-based} Regular grid-based representations~\cite{hui2022neural, hui2024make,li2023diffusion} utilize a fixed-resolution grid for encoding various 3D quantities such as occupancy, signed distance function (SDF), color and etc. Such methods, however, may not capture extremely fine structures at the very initial stage of discretization. Additionally, regular grid-based representations allocate a substantial portion of their capacity to the empty spaces, leading to a potential cubic rate of parameter increase. In contrast, the adaptive local grids in \acro are able to capture thin structures and the octree forest allows for scaling the representation power with only a linear increase of parameters. 
Deep Marching Tetrahedra~\cite{shen2021deep} (DMTet) uses deformable tetrahedral grids to improve the representation efficiency compared with regular grids. However, the fitting time~\cite{liu2023meshdiffusion} of DMTet is about 10 times ours, the parameter count of DMTet is 3 times ours (See~\autoref{tab:compare_property} for specific numbers.) while we represent geometry more accurately than it. 


\textbf{Triplane} To mitigate the cubic complexity inherent in regular grid-based methods, triplane approaches~\cite{chan2022efficient,shue20233d,gupta20233dgen} employ three planes, each aligned with an axis in 3D space. Information at each 3D location is queried by first tri-linear interpolation from the three planes, followed by processing through various neural network architectures. Although triplane-based representations enhance parameter efficiency compared to regular grid-based methods, the tri-linear interpolation may result in oversmoothed outcomes and also missing geometric details.

\textbf{Point Cloud} Point cloud-based representations ~\cite{luo2021diffusion, zhou20213d, zeng2022lion} describe 3D shapes by placing points on geometry boundaries. To accurately describe the shape, excessive number of points will be needed, which is not practical for the following generation task. Limited number of points will loss details of the shape. Moreover, post-processing methods~\cite{kazhdan2006poisson,peng2021shape} are required to convert point clouds to meshes, which may introduce additional inductive bias.

\textbf{Neural Implicit} Neural implicit representations~\cite{park2019deepsdf,chen2019learning} model 3D shapes by learning mappings between coordinates and desired 3D quantities. When combined with the autoencoding framework, latent codes~\cite{zhang20233dshape2vecset,chou2023diffusion} are derived. A drawback of these methods is that the autoencoding may result in oversmoothed geometric features. Furthermore, mesh extraction from neural implicit representations requires querying numerous points in space, yielding expensive neural network evaluations. In contrast, \acro enables direct and efficient extraction of geometries without the involvement of any neural network. 

Lastly, we would address the work of Mosaic-SDF~\cite{yariv2023mosaic} individually. Different from our method, Mosaic-SDF only has a single level of voxels and each voxel has a isotropic scale for each axis. In this sense, it has limited adaptation in representing the geometry. Moreover, it has to model the whole distribution as a whole at once while previous researches~\cite{saharia2022photorealistic,pernias2023wuerstchen} have shown hierarchical or cascaded generation may benefit the generation.


\subsection{Graph and Tree Generation}

General graph generation relies on the Graph Neural Network and message passing mechanism. Others like HGGT~\cite{jang2023hierarchical} use graph compression methods such as $K^2-$Tree and sequentially generate tree nodes in an autoregressive way. In our case, generating an arbitrary topology graph is unnecessary because we focus on generating octrees. OGN~\cite{tatarchenko2017octree} uses recursive way of generating octree voxels and Octree Transformer~\cite{ibing2023octree} uses autoregressive way of generating tree nodes by transformers. In contrast with their methods, we use the widely adopted and effective generation paradigm, the diffusion model. BrepGen~\cite{xu2024brepgen} generates fixed topology face-edge-vertice trees hierarchically by diffusion. Since the topology is fixed, the corresponding relation can be easily injected when training the diffusion model. We plan to adopt similar cascaded generation scheme to BrepGen based on our \acro representation.

\section{Method}
We will introduce our methods in two parts: \autoref{sec:construction_of_OFALG} the construction of \acro from the input watertight meshes and \autoref{sec:cascaded_diffusion_of_OFALG} the cascaded generation of \acro.

\begin{figure}[th]
    \centering

\includegraphics[width=0.8\textwidth]{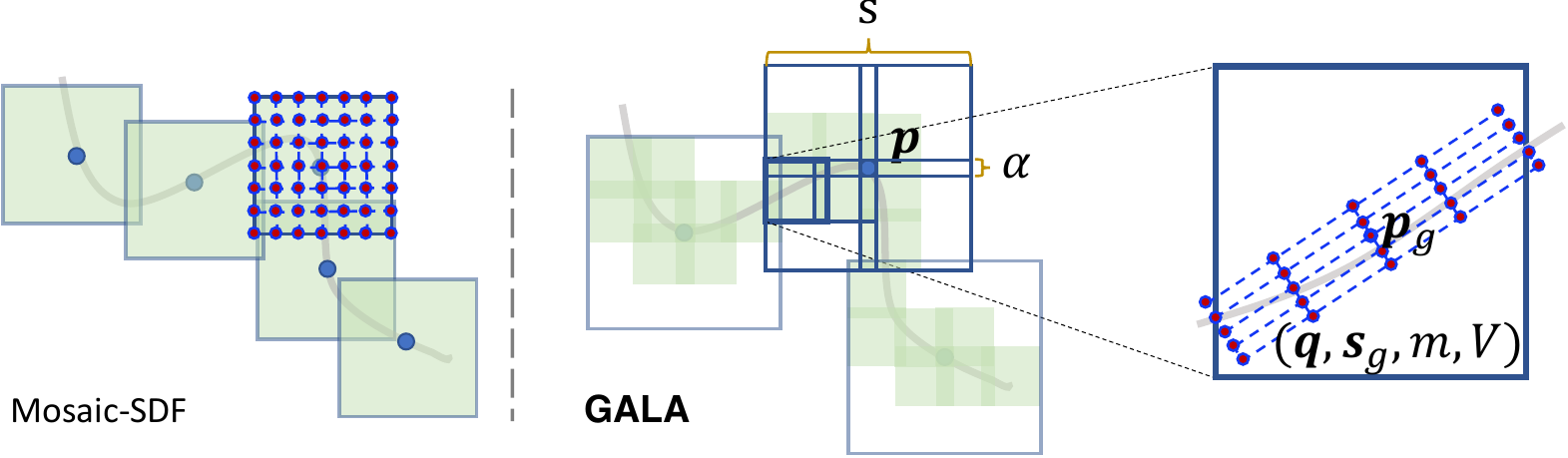}
    \caption{In our representation, \textbf{\acro}, tree root nodes, as voxels, are initialized over mesh surfaces (\textbf{\textcolor[rgb]{0.843 0.835 0.835}{gray line}}), each with location $\mathbf{p}\in\mathbb{R}^3$ and scale $s\in\mathbb{R}$. Descendant node voxels are deduced recursively with each child voxel, with overlapping, and expanded at ratio $\alpha\in\mathbb{R}$ into depth $d$. Only at a {\em non-empty\/} leaf node subdivision (\textbf{\textcolor[rgb]{0.7725, 0.8784, 0.70588}{light green}}) would a local adaptive grid  of resolution $m\in\mathbb{N^{+}}$ be extracted with location $\mathbf{p}_g\in\mathbb{R}^3$, orientation quaternion $\mathbf{q}\in\mathbb{R}^4$ , scales $\mathbf{s_g}\in\mathbb{R}^3$, and values $V\in\mathbb{R}^{m^3}$. $N_o,\alpha,m,d$ are hyperparameters. In contrast, Mosaic-SDF~\cite{yariv2023mosaic} (left) employs mosaic patches, each {\em fully\/} occupied by a {\em single-level\/}, {\em axis-aligned\/}, and {\em isotropic\/} grid.}
    \label{fig:overall_OFALG}
\end{figure}

\subsection{Construction of \acro}
\label{sec:construction_of_OFALG}

We illustrate the overall \acro structure in 2D in~\autoref{fig:overall_OFALG} and will dissect the construction process into:~\autoref{sec:octree_forest_init} Octree forest initialization,~\autoref{sec:anisotropic_local_grid_extraction} Local adaptive grid extraction,~\autoref{sec:refinement} Refinement and~\autoref{sec:quantization} Fitting-aware Quantization.

\subsubsection{Octree Forest Initialization}
\label{sec:octree_forest_init}
Given the number of trees $N_o$, we find the centroids $\mathbf{p}$ of tree root node voxels by Farthest Point Sampling (FPS)~\cite{eldar1997farthest} of densely sampled points $\mathcal{X}$ on the mesh of the input object. Then, to determine the scale $s$ of each root voxel $i$, we assign each sampled point $\mathbf{x} \in \mathcal{X}$ to its nearest centroid and define the scale scalar as the infinity-norm radius of each cluster:

\begin{equation}
s_i = \max_{\mathbf{x} \in \mathcal{C}_i} \| \mathbf{x} - \mathbf{p}_i \|_{\infty}, \quad \mathcal{C}_i = \left\{ \mathbf{x} \in \mathcal{X} \ \bigg| \ i = \underset{k \in \{0, \dots, N_o - 1\}}{\arg \min} \| \mathbf{x} - \mathbf{p}_k \|_2 \right\}
\end{equation}

 In this way, we ensure that the forest will fully cover the shape boundaries. After setting the tree root voxels, we recursively deduce the further levels of tree nodes in an overlapping way where we slightly expand the divided spaces at each level by ratio $\alpha$. This expansion operation enables more voxels to contribute to the geometry modeling at early 
 tree levels. See \ref{sec:ablation_study_of_overlapping_subdivison} for the ablation study of $\alpha$.

\subsubsection{Local Adaptive Grid Extraction}
\label{sec:anisotropic_local_grid_extraction}
\vspace{-8pt}

\begin{figure}[th]
    \centering
    \includegraphics[width=0.9\textwidth]{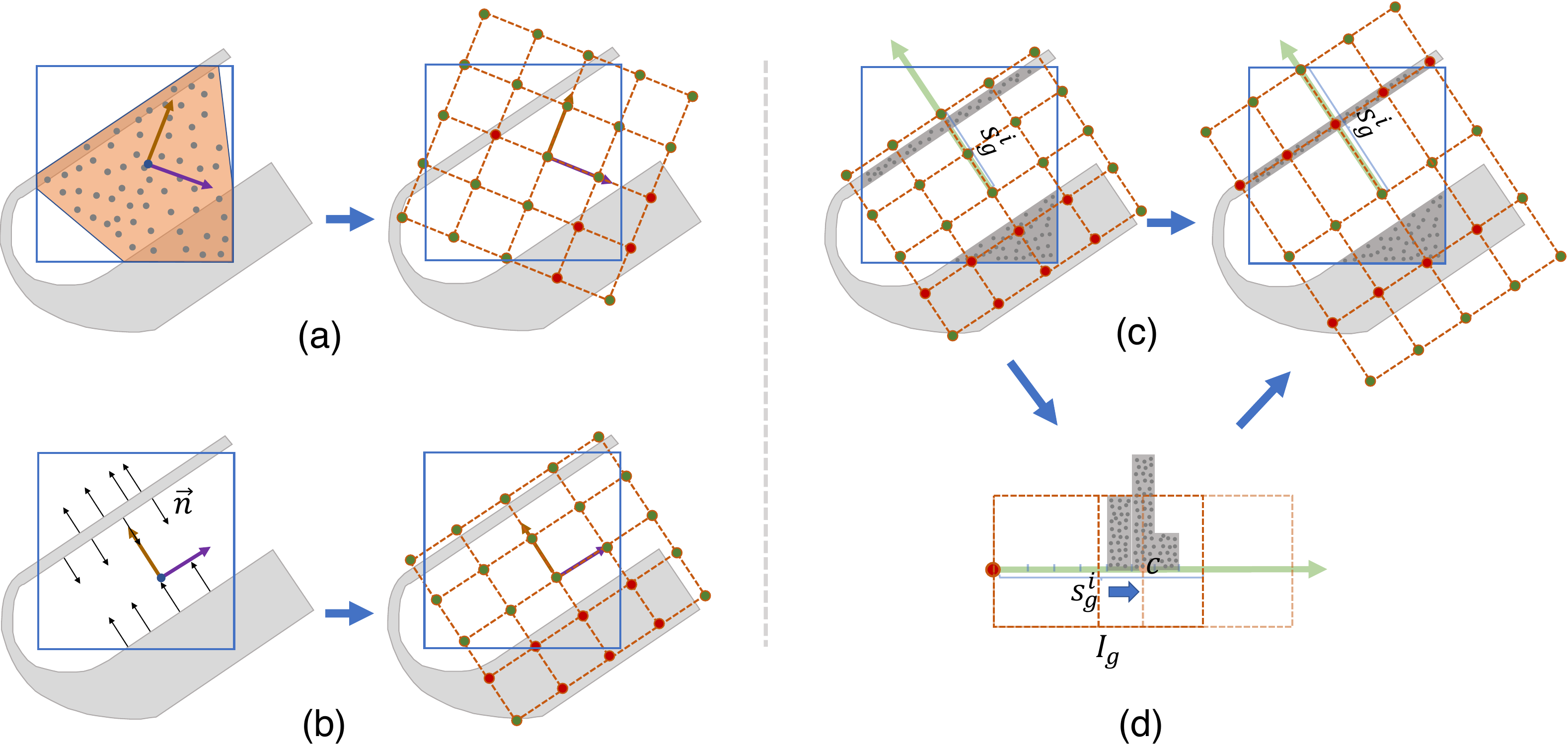}
    \caption{Illustration of local adaptive grid extraction in 2D. Within each subdivision (\textbf{\textcolor[rgb]{0.2667,0.447059,0.768627}{blue square}}), \textbf{(a)} an OBBTree~\cite{gottschalk1996obbtree} determines the orientation of the local bounding box according to the \textbf{\textcolor[rgb]{0.929, 0.753, 0.620}{convex hull}} of the subdivided \textbf{\textcolor[rgb]{0.843 0.835 0.835}{local geometry}}. \textbf{(b)} Different from OBBTree, we determine the orientation of the \textbf{\textcolor[rgb]{0.7725490,0.3529411,0.0666666}{local adaptive grid}} using PCA on bounded normal vectors ($\overrightarrow{n}$). \textbf{(c)} Moreover, we rescale the grid anisotropically to better capture the local geometry, as informed by the histogram \textbf{(d)} of sampled points (\textbf{\textcolor[rgb]{0.498 0.498 0.498}{small dark gray dots}}) on triangle meshes projected along each \textbf{\textcolor[rgb]{0.7215686, 0.839215, 0.650980}{axis}} of the grid. The \textcolor[rgb]{0.75294,0,0}{\textbullet}/\textcolor[rgb]{0.32941,0.50980,0.20784}{\textbullet} are grid samples with negative/positive signs.
}
    \label{fig:grid_extraction}
\end{figure}


Properly placing the local adaptive grid within each subdivision is essential to our method. The OBBTree~\cite{gottschalk1996obbtree} determines the orientation of the local bounding box using eigenvectors of the covariance matrix derived from points sampled on the convex hull faces of the subdivided geometry (Figure~\ref{fig:grid_extraction}(a)). However, we take a different approach. Instead of only defining a bounding box, we use a \textbf{grid} designed to capture as much geometric information as possible. Because the orientation derived from convex hull sample points may not align with the actual geometric structure, we align the grid orientation with PCA of the local normal vectors $\overrightarrow{n}$ instead (Figure~\ref{fig:grid_extraction}(b)).

Additionally, we rescale the grid so that at least one set of grid samples locates at the most geometrically dense spot and adjacent grid samples with different signs capture more zero isosurfaces (Figure~\ref{fig:grid_extraction}(c)).  To achieve this, we utilize the histogram $\mathcal{H}$ of sampled points projected along each grid axis $\mathbf{u}_i$ and move nearest and smaller grid samples to the histogram peak (Figure~\ref{fig:grid_extraction}(d), Algorithm~\ref{alg:rescaling_grid_with_histogram})


\begin{algorithm}
\caption{Rescaling the Grid with Histogram}
\label{alg:rescaling_grid_with_histogram}
\begin{minipage}{.48\textwidth}
\begin{algorithmic}[1]
\State \text{Input: orientation $\mathbf{u}_i$ and scale $s_g^i$ at axis $i$;} 
\Statex \text{center $\mathbf{p}_g$ and bounded meshes $\mathcal{M}$}
\State Split half axis of length $s_g^i$ into $n_h$ bins
\State Sample points $\mathbf{x}\in\mathcal{M}$
\State Count \text{$|\frac{(\mathbf{x}-\mathbf{p}_g)\cdot\mathbf{u}_i}{\|(\mathbf{x}-\mathbf{p}_g)\|}|$} into histogram $\mathcal{H}$
\end{algorithmic}
\end{minipage}%
\begin{minipage}{.48\textwidth}
\begin{algorithmic}[1] 
\setcounter{ALG@line}{4} 
\State $c \gets \text{center of } \max(\mathcal{H})$
\State Get grid index $I_g \gets \arg\max_I \left(\frac{2I}{m} < c\right)$
\If{$I_g > 0$}
    \State $s^i_g \gets \frac{m}{2I_g \cdot c}$
\EndIf
\end{algorithmic}
\end{minipage}
\end{algorithm}

We will show the effectiveness of our local adaptive grid extraction by ablation studies in~\autoref{sec:ablation_study_of_anisotropic_local_grid_extraction}. After determining the locations, orientations and scales of anisotropic local grids, truncated ground truth SDF~\cite{hui2022neural} values (truncated to $[-0.1, 0.1]$) are used to initialize the grid values $V$. All the above initialization and extraction processes before refinement cost few seconds (as $2.56\pm1.30$s in ShapeNet Airplane). More detailed information on the time measurement, see \ref{subsec:time_measurement}.

\subsubsection{Refinement}
\label{sec:refinement}
After extracting the local adaptive grids, we further refine the grid values $V$ by backpropagation to represent the input geometry more accurately. For each queried location $\mathbf{x}$, the queried value $v_{\mathbf{x}}$:



\begin{equation}
v_{\mathbf{x}} = \frac{1}{\sum_i w(\mathbf{x},i)}\sum_{i}w(\mathbf{x},i) \mathbf{I}(\mathbf{x}, i),
\label{equ:extract}
\end{equation}
\begin{equation}
    w(\mathbf{x},i)=\text{ReLU}\big[1-\|(\mathbf{x}-\mathbf{p}_g^i)\mathbf{O}_i\cdot \mathbf{s}_i^{-1}\|_\infty\big],
    \label{equ:weight}
\end{equation}

where $\mathbf{I}(\mathbf{x},i)$ (\autoref{equ:extract}) is the tri-linear interpolation of grid values $V$ within the anisotropic local grid $i$ that encapsulates the query $\mathbf{x}$ and $w(\mathbf{x},i)$ (\autoref{equ:weight}) is the weighting function which is defined by the infinity ball centered at the grid center $\mathbf{p}_g^i$ and rotation matrix $\mathbf{O}_i\in\mathbb{R}^{3\times3}$. All grid values will be optimized by MSE loss $L_{\text{MSE}} = \mathbb{E}(\|v_{\mathbf{x}} - v_{\text{\tiny GT}}(\mathbf{x})\|_2)$ using backpropagation, where $v_{\text{\tiny GT}}(\mathbf{x})$ is the truncated ground truth SDF value at location $\mathbf{x}$ directly computed from input meshes. We assign $v_{\mathbf{x}}=0.1$ if $\mathbf{x}$ does not lie in any local grid. We will flip these signs of interior parts to $-0.1$ via a DFS method detailed in \ref{sec:flip_sign} when extracting meshes from \acro.

\subsubsection{Fitting-aware Quantization}
\label{sec:quantization}
To increase the storage (see more at \ref{sec:storage_efficiency}) and memory efficiency of \acro as well as lower the difficulty of aligning float values in space inspired by the discrezation step in MeshGPT~\cite{siddiqui2024meshgpt}, we quantize the local adaptive grid related quantities $\mathbf{q},\mathbf{p}_g,\mathbf{s_g},V$ during the fitting process by Straight-Through Estimator (STE)~\cite{bengio2013estimating,han2015deep}. We quantize the orientation in the unit of $\frac{\pi}{60}$ of Euler angles and $\mathbf{s}_g,\mathbf{p}_g,V$ in 8 bit of the range $[0,0.1],[-0.5, 0.5], [-0.1,0.1]$. 


\subsection{Cascaded Generation of \acro}
\label{sec:cascaded_diffusion_of_OFALG}

Motivated by previous cascaded generation works~\cite{saharia2022photorealistic,ho2022cascaded,xu2024brepgen,hui2022neural,chen2021decor} which successfully generate complicated 2D or 3D data in a coarse-to-fine manner, we generate sets of information $X_o,X_V,X_{\bar{V}}$ of \acro step by step where flattened vectors of root node $\mathbf{x}_{o}\in\mathbb{R}^4=\mathbb{R}^3$($\mathbf{p}$) $+\mathbb{R}$($s$), grid values
 $\mathbf{x}_{V}\in\mathbb{R}^{m^3}$, and grid configuration $\mathbf{x}_{\bar{V}}\in\mathbb{R}^{10}=\mathbb{R}^4$($\mathbf{q}$)$+\mathbb{R}^3$($\mathbf{s}_g$)$+\mathbb{R}^3$($\mathbf{p}_g$) belong to one of the sets respectively (Refer to \ref{sec:data_processing} for more details on the data preparation). Specially, since the configurations of local adaptive grids $X_{\bar{V}}$ already encode rich local geometric information as articulated in~\autoref{sec:anisotropic_local_grid_extraction}, we are able to directly regress $X_V$ from $X_{\bar{V}}$, which greatly alleviates the generation training burden.



\begin{figure}[th]
    \centering
    \includegraphics[width=1.0\textwidth]{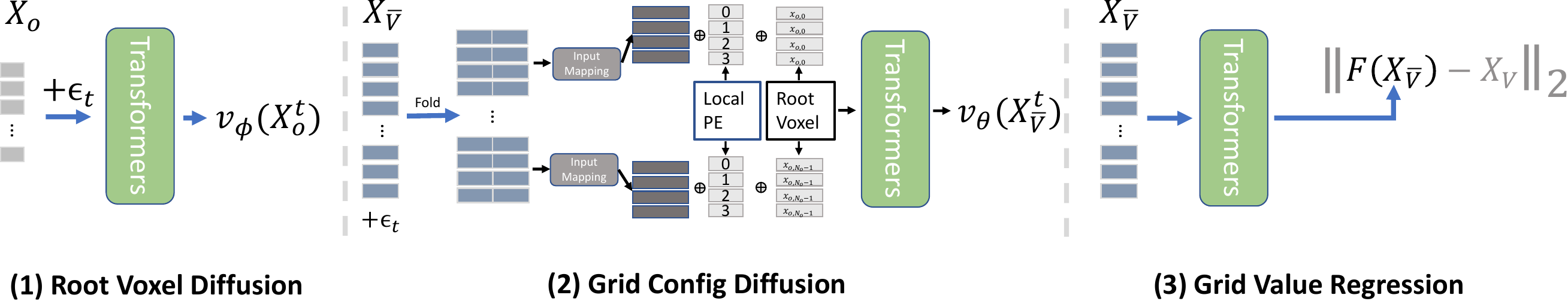}
    \caption{\acro generates shapes in a cascaded manner by (1) Root voxel diffusion; (2) Local adaptive grid configuration diffusion; (3) Local adaptive grid value prediction by regression.}
    \label{fig:cascaded_diffusion_overall}
\end{figure}

  All sets of vectors are processed by transformer-based networks~\cite{vaswani2017attention,peebles2023scalable}. (1) The class label $y$ conditioned distribution of root voxels $P(X_o|y)$ is firstly modeled by diffusion model~\cite{song2020score,ho2020denoising}. (2) Configurations of anisotropic local grids $X_{\bar{V}}$ are generated following that via modeling $P(X_{\bar{V}}|X_o,y)$ by diffusion model. (3) Lastly, grid values $V$ are predicated from grid configurations $\bar{V}$ via $X_V = F(X_{\bar{V}},y)$ and trained by L2 regression loss. More specifically speaking, as we can see in~\autoref{fig:cascaded_diffusion_overall}.(2), siblings of padded leaves are grouped and folded once and previously generated root voxel information $x_{o,i}\in X_o$ is added directly to the nodes of its corresponding descendants together with local positional encodings (PE) of trees to differentiate siblings among them.  In all, following the popular diffusion training paradigm~\cite{ho2020denoising,song2020score}, we update the diffusion network weights $\theta$ via the v-prediction~\cite{salimans2022progressive}:
\begin{equation}
    \nabla_{\theta}\|v - v_{\theta}\big(X^t_{\bar{V}},t,y,X_o\big)\|^2,\quad v=\alpha_t X^0_{\bar{V}}-\sigma_t\epsilon
\end{equation}

where $X^t_{\bar{V}}=\sqrt{\bar{\alpha_t}}X^0_{\bar{V}} + \sqrt{1-\bar{\alpha_t}}\epsilon, X^0_{\bar{V}}\sim q(X_{\bar{V}}), \epsilon \sim \mathcal{N}(\mathbf{0},\mathbf{I})$ is the diffusion forward process. $\bar{\alpha}_t$ is defined by $\bar{\alpha}_t=\prod^t\alpha_t,\alpha_t=1-\beta_t$ where $\beta_t$ is the variance schedule. $\sigma_t = \sqrt{1 - \bar{\alpha}_t}$ and cosine noise scheduling~\cite{nichol2021improved} is used.  The network weights $\phi$ for the first step are updated in a similar manner as described above. For more details, please refer to \ref{sec:more_details_on_generation}.


\section{Experiments and Results}

\subsection{Implementation Details}
For the hyperparameter of \acro, we set tree root number $N_o=256$, child node expanding ratio $\alpha=0.2$, grid resolution $m=5$ and depth $d=1$. The number of histogram bins $\mathcal{H}$ is $2m$. For the refinement process, 8192 points are queried near surfaces on each run for 400 iterations.  Details of the pure C++/CUDA implementation will be found at \ref{sec:details_of_the_pure_implementation}.

For the cascaded generation, each network for every generation step is consisted of 24 transformer layers with model channel 1024. AdamW~\cite{loshchilov2017decoupled} with $\text{lr}=8e^{-5},\beta_1=0.9,\beta_2=0.999$ is adopted. We test the reconstruction and generation ability of our methods on the whole ShapeNet~\cite{chang2015shapenet} dataset. We follow the dataset split of previous works~\cite{zhang20223dilg,zhang20233dshape2vecset,yariv2023mosaic} and train conditional generation on the whole ShapeNet classes.
 

\subsection{Metrics}

We evaluate reconstruction results using Chamfer Distance (CD) between two sampled 3D objects and also report the Light Field Distance (LFD)~\cite{chen2003visual}, a widely adopted~\cite{chen2020bsp,gao2022get3d,zuo2023dg3d} visual similarity metric that is well-known from computer graphics. To evaluate the generation results, we adopt the metric of Maximum Mean Discrepancy (MMD), Coverage (COV), and 1-nearest neighbor accuracy (1-NNA) based on Chamfer Distance (CD) and Earth Mover Distance (EMD)~\cite{luo2021diffusion,zeng2022lion}.


\subsection{Reconstruction Results}
\label{sec:recon}

\begin{figure}[th]
    \centering
    \includegraphics[width=1.0\textwidth]{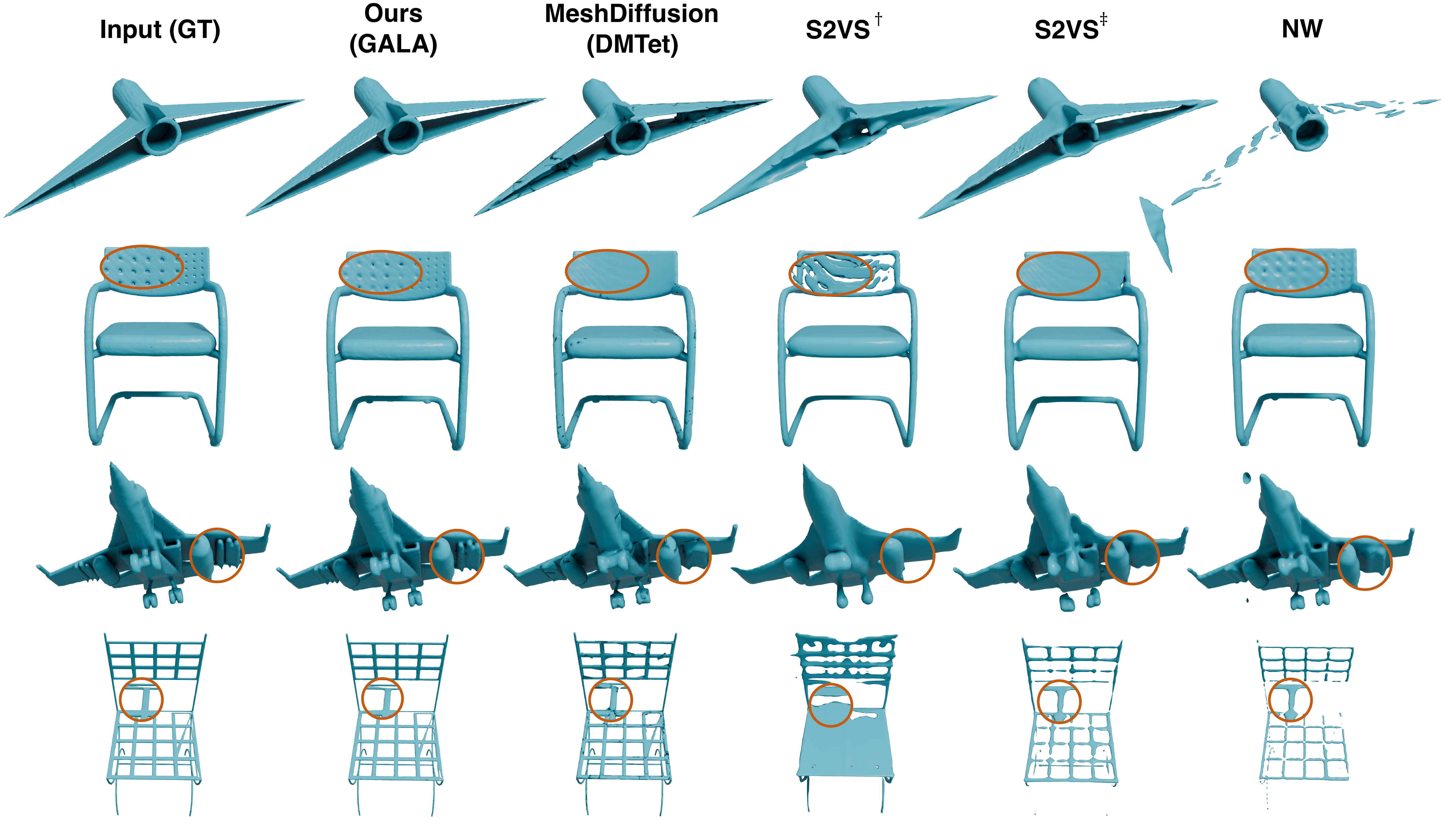}
    \caption{Qualitative comparison of reconstruction results between ours and other important baselines. S2VS$^\dagger$~\cite{zhang20233dshape2vecset} is evaluated from the public checkpoint provided in the official repository and S2VS$^\ddagger$ is our own implementation. All evaluated at resolution $256^3$. Zoom in to see more details. }    \label{fig:recon_comparison}
\end{figure}

We compare our reconstruction results with 3DShape2VecSet (S2VS)~\cite{zhang20233dshape2vecset}, Neural Wavelet (NW)~\cite{hui2022neural} and MeshDiffusion (DMTet, $64^3\times 4$)~\cite{liu2023meshdiffusion}. We measure the reconstruction metrics on 50 objects of ShapeNet classes (airplane, car, char, basket, guitar), 10 objects per class. From \autoref{tab:recon_results}, we can see that our reconstruction is more accurate than other baselines. $\mathcal{A}$ means local adaptive grids. Without $\mathcal{A}$, uniform grid samplings will be placed within tree voxels in the axis-aligned way. Visual comparisons on results w/o $\mathcal{A}$ are shown in \ref{sec:more_on_wo_A}, where the local adaptive grid setting captures more accurate details on geometry surfaces. Qualitative comparison of reconstruction is shown in~\autoref{fig:recon_comparison}. As we can see from~\autoref{fig:recon_comparison}, we preserve the most geometric details with one of the fewest parameter numbers.


\begin{table}[ht]
\centering
\begin{tabular}{lccccc} 
\toprule
Metric &   S2VS&
NW&
MeshDiffusion&
Ours, w/o $\mathcal{A}$ &
Ours \\
\midrule
LFD $\downarrow$  & 2429.92 & 1869.80 & 1301.14 & 1146.74 & \cellcolor{lightblue} 1109.36 \\
CD $\downarrow$ ($\times10^{-3}$) & 4.32 & 4.17 & 2.61 & 1.52 & \cellcolor{lightblue} 1.18 \\   \#Parameters & $\leq$ 262k & 536k & 1.05M & 263k & 277k \\
\bottomrule
\end{tabular}
\caption{Reconstruction quality across 5 ShapeNet classes (airplane, basket, car, chair, guitar; 10 random objects per class).}
\label{tab:recon_results}

\end{table}

\begin{figure}[ht]
    \centering
    \includegraphics[width=0.95\textwidth]{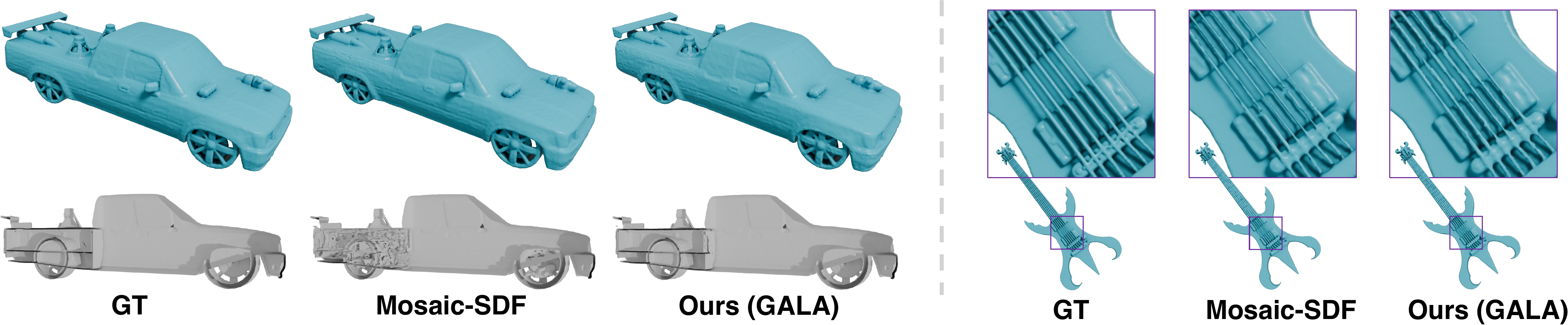}
    \caption{Qualitative comparison of the reconstruction result with Mosaic-SDF~\cite{yariv2023mosaic}. Mosaic-SDF results are provided by the author directly. We reveal the complicated internal structures of the input GT and the reconstructed results by slicing them based on  Slice3D~\cite{wang2023slice3d}.}
    \label{fig:MSDF_comparison}
\end{figure}

Because we do not have the official code release from Mosaic-SDF~\cite{yariv2023mosaic}, we requested the reconstruction results from the authors of Mosaic-SDF and compare the reconstruction of ours with them qualitatively in~\autoref{fig:MSDF_comparison}. As we can see from~\autoref{fig:MSDF_comparison}, the car is a challenging example shown in Mosaic-SDF and with complicated exterior and interior structures such as very thin and close surfaces. \acro can capture exterior and interior geometry structures at the same time. In contrast, the result of Mosaic-SDF fails to capture internal structures. As for the guitar example, \acro has the better connection of guitar strings and the thickness of strings is closer to the ground truth than it of Mosaic-SDF.

\subsection{Generation Results}

\begin{table}[th]
\centering
\resizebox{\textwidth}{!}{
\begin{tabular}{lcccccc|cccccc}
\toprule
& \multicolumn{6}{c}{\itshape airplane} &
\multicolumn{6}{c}{\itshape chair} \\
\cmidrule{2-13}
 & \multicolumn{2}{c}{COV($\uparrow$,\%)} & \multicolumn{2}{c}{MMD($\downarrow$)} & \multicolumn{2}{c|}{1-NNA($\downarrow$,\%)} & \multicolumn{2}{c}{COV($\uparrow$,\%)} & \multicolumn{2}{c}{MMD($\downarrow$)} & \multicolumn{2}{c}{1-NNA($\downarrow$,\%)}\\
 & CD & EMD & CD & EMD & CD & EMD & CD & EMD & CD & EMD & CD & EMD \\
\midrule
3DIGL& 41.09 & 32.67 & 4.69 & 4.73 & 82.67 & 84.41 & 
37.87 & 39.94 & 20.37 & 10.54 & 74.11 & 69.38 \\
NW& 51.98 & 45.05 & \cellcolor{lightgray}3.36 & 4.19 & 68.32 & 73.76 &
43.79 & 47.04 & 16.53 & 9.47 & 59.47 & 64.20 \\
S2VS& 51.98 & 40.59 & 3.80 & 4.45 & 69.06 & 76.73 &  
\cellcolor{lightgray}51.78 & 52.37 & 16.97 & 9.44 & 58.43 & 60.80 \\
Mosaic-SDF& \cellcolor{lightgray}52.48 & \cellcolor{lightgray}51.49 & 3.54 & \cellcolor{lightgray}3.78 & \cellcolor{lightgray}62.62 & \cellcolor{lightgray}69.55 &
48.22 & \cellcolor{lightblue}55.03 & \cellcolor{lightgray}15.47 & \cellcolor{lightgray}9.13 & \cellcolor{lightblue}51.04 & \cellcolor{lightgray}55.62 \\
Ours    & \cellcolor{lightblue} 52.91 & \cellcolor{lightblue} 51.49 & \cellcolor{lightblue}2.48 & \cellcolor{lightblue}2.84 & \cellcolor{lightblue}59.90 & \cellcolor{lightblue}57.43 & \cellcolor{lightblue}52.21 & \cellcolor{lightgray}53.10 & \cellcolor{lightblue}13.28 & \cellcolor{lightblue}8.17 & \cellcolor{lightgray}53.10 & \cellcolor{lightblue}54.57 \\
\bottomrule
\end{tabular}
} 
\caption{Generation quality for ShapeNet classes of airplane and chair. By convention, MMD-CD has been multiplied by $10^3$ and MMD-EMD by $10^2$ for readability.}
\label{tab:generation_results}
\end{table}

\begin{figure}[htbp]
    \centering
    \begin{subfigure}[b]{1.0\textwidth}
        \centering
        \includegraphics[width=\textwidth]{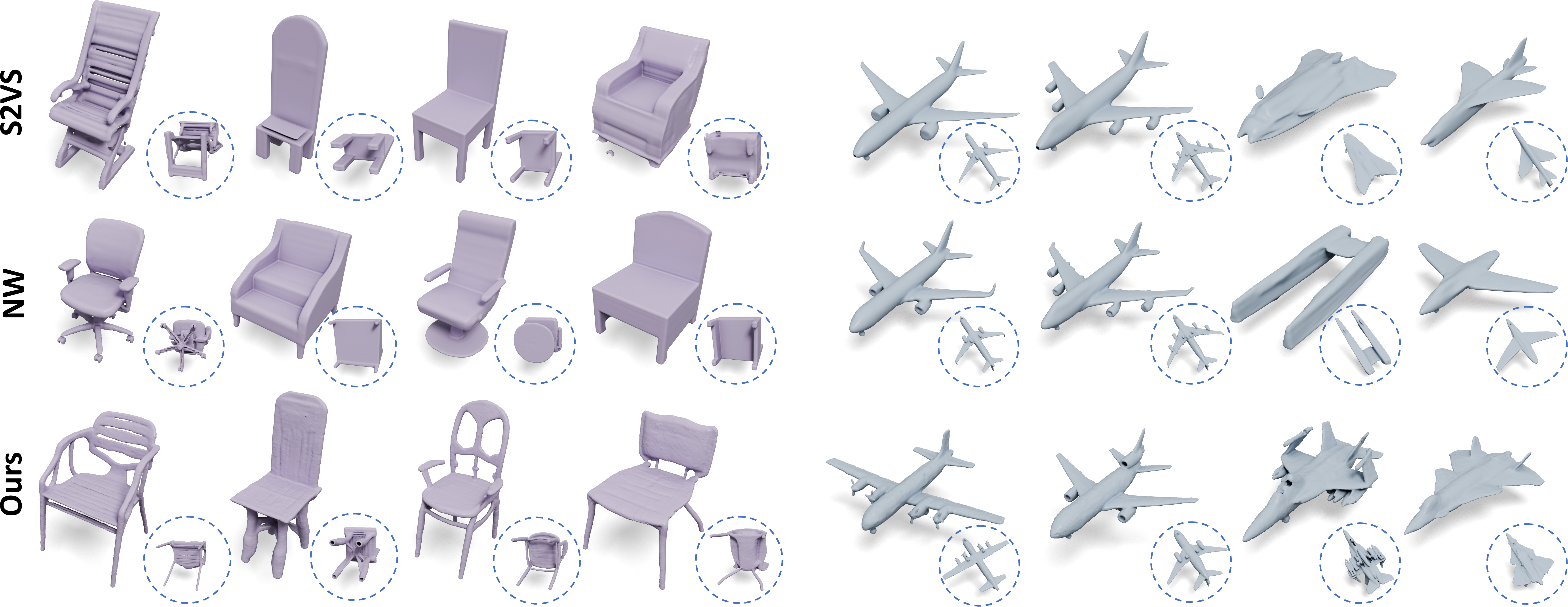}
    \end{subfigure}

    \vspace{2pt} 

    \begin{subfigure}[b]{1.0\textwidth}
        \centering
        \includegraphics[width=\textwidth]{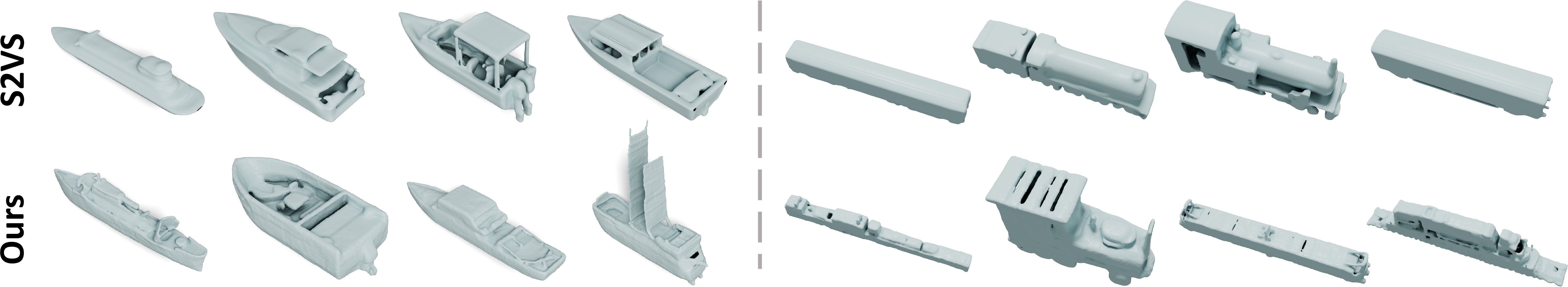}
    \end{subfigure}

    \caption{The qualitative comparison of generation results. Additional results of ShapeNet vessel and train are shown between S2VS and ours to demonstrate the detailed generation of ours.}
    \label{fig:generation_comparison}
\end{figure}

From \autoref{tab:generation_results}, we show our cascaded generation on \acro produces shapes better than other baselines quantitatively. We also provide qualitative results on \autoref{fig:generation_comparison} to show that our generation gives much geometry details in generation results. Please see the supplementary file for more generation results.

\vspace{-10pt}

%
%

\subsection{Ablation Studies}

\paragraph{Local Grid Extraction.}
\label{sec:ablation_study_of_anisotropic_local_grid_extraction}

\begin{wraptable}[10]{r}{0.5\textwidth}
\vspace{-5pt}
\centering
\begin{tabular}{@{}lccc} 
\thickhline
\addlinespace[4pt] 
Metrics & vanilla &   + $\overrightarrow{n}$ & + $\overrightarrow{n},\mathcal{H}$ \\
\midrule
CD $\downarrow$ ($\times10^{-3}$)    &  1.55  
                        &  1.33 & 1.18 (1.08$^\dagger$) \\
HD $\downarrow$ ($\times10^{-2}$)   &  1.46 
                       &  1.19 & 1.06 (0.98$^\dagger$) \\
\addlinespace[4pt] 
\thickhline
\end{tabular}
\vspace{2pt}
\caption{Ablating the anisotropic local grid extraction components ($\overrightarrow{n},\mathcal{H}$) on the same set of data of reconstruction test. $^\dagger$: without quantization.}

\label{tab:grid_extraction_ablation}
\end{wraptable}

\begin{figure}[ht] 
    \centering
    \includegraphics[width=1.0\textwidth]{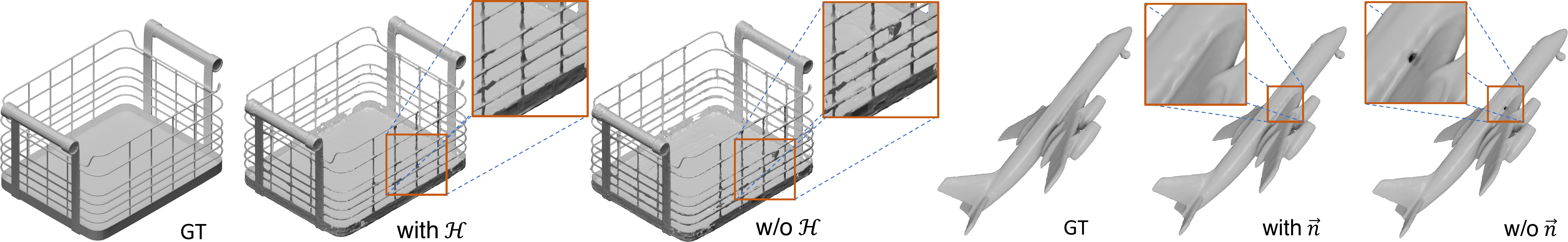}
    \caption{Qualitative examples of \acro fitting results on \textbf{Left}: rescaling the local grid with histogram or not; \textbf{Right}: determining the grid orientation with normals or not. Zoom in to see more.}
    \label{fig:grid_extraction_ablation}
\end{figure}

We conduct the ablation study on the components in local adaptive grid extraction, including determining orientation by bounded normals ($\overrightarrow{n}$) and rescaling with histogram ($\mathcal{H}$). \autoref{tab:grid_extraction_ablation} shows the effectiveness of our local adaptive grid extraction design and limited degradation by the quantization procedure. We further give some qualitative examples in \autoref{fig:grid_extraction_ablation} to show that corresponding components help better capture the detailed geometries.

\begin{figure}[!ht]
  \centering
  \includegraphics[width=1.0\textwidth]{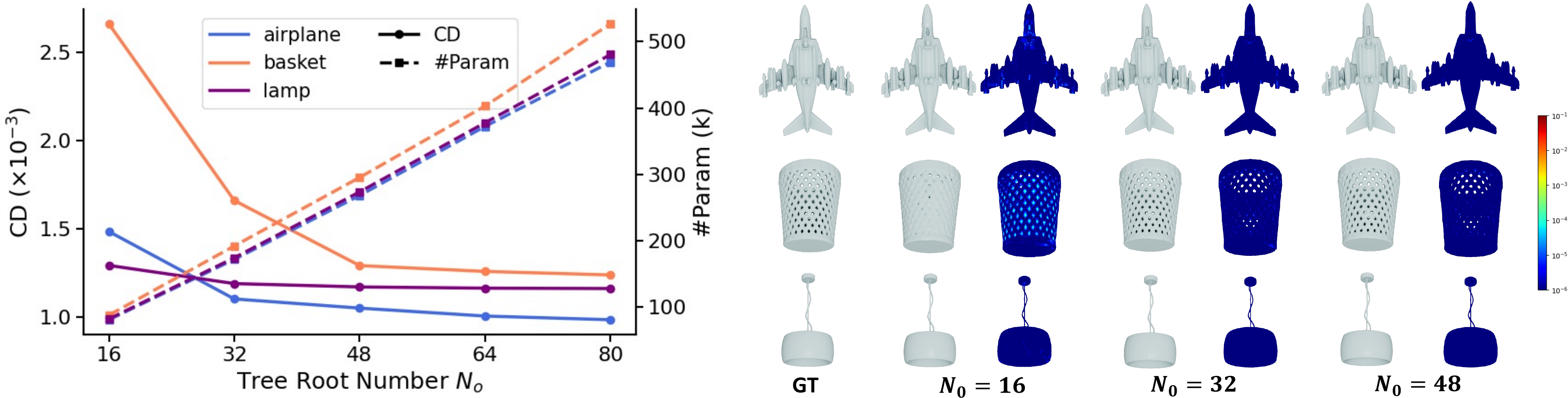} 
  \caption{Influence of tree count $N_o$ (with depth $d=2$) on three examples.}
  \label{fig:more_tree_setting}
\end{figure}

\paragraph{More Tree Setting.}
%
We examine the effect of varying tree counts $N_o$ (with depth $d = 2$) on three examples, as shown in~\autoref{fig:more_tree_setting}. As we can see in~\autoref{fig:more_tree_setting}, using $N_o=32$ with fewer than 200k parameters already yields low-error reconstructions across the three distinct example objects. Furthermore, even at $N_o=16$ and fewer than 100k parameters, our method effectively captures significant geometric details, such as the missiles on the airplane's rear and the lamp's strings, demonstrating the method's great adaptability.

\vspace{-10pt}

\paragraph{Overlapping Subdivision}
\label{sec:ablation_study_of_overlapping_subdivison}
\begin{wrapfigure}[7]{r}{0.4\textwidth} 
  \centering
  \includegraphics[width=0.4\textwidth]{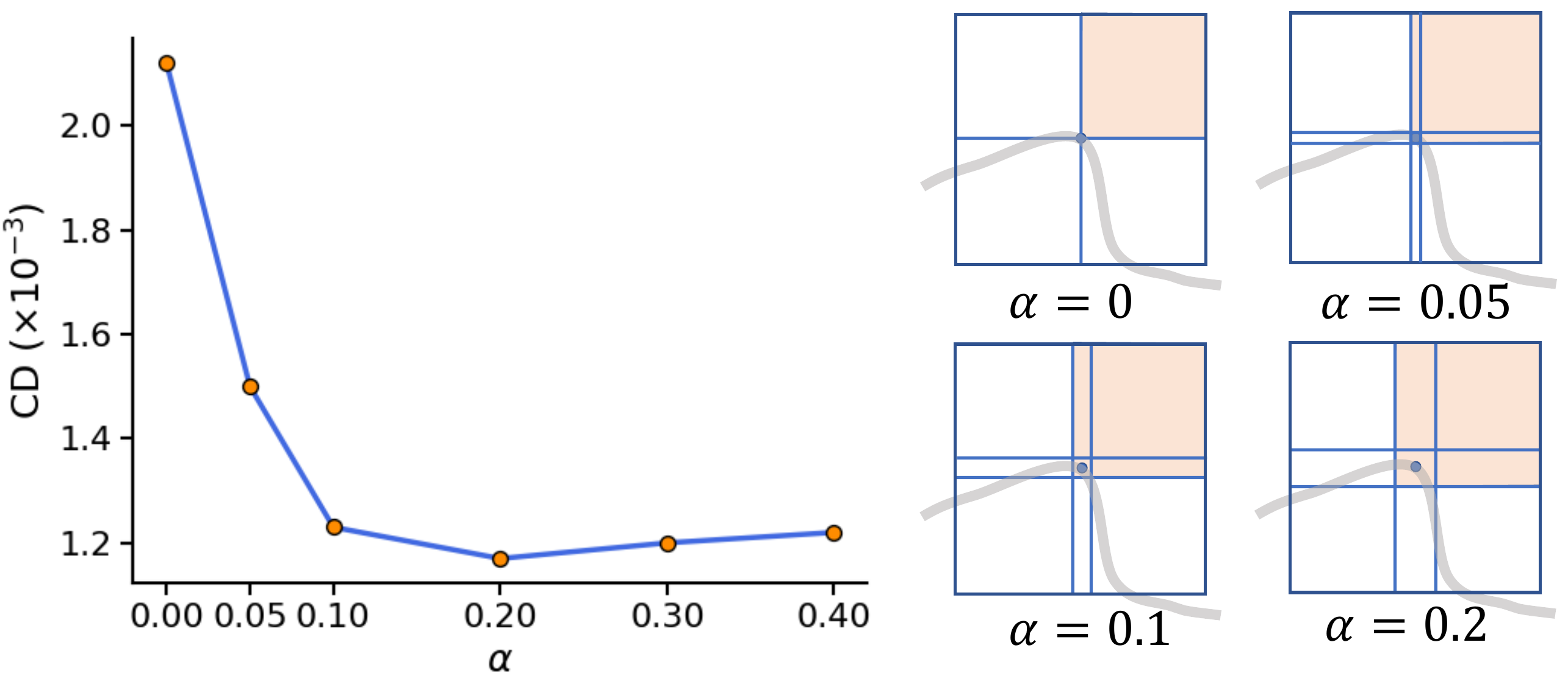} 
  \captionsetup{font=scriptsize}
  \caption{Ablation of child overlap $\alpha$.}
  \label{fig:alpha_ablation}
\end{wrapfigure}

We show the reconstruction performance by presenting Chamfer Distance (CD) of different overlap ratios $\alpha$ of subdivision in~\autoref{fig:alpha_ablation}. To avoid making the tree too sparse at early levels while maintaining the tree structures, we expand the subdivision with ratio $\alpha$. However, expanding too much would decrease the granularity of each subdivision, we find a sweet spot $\alpha=0.2$ by scanning it on the reconstruction test set.

\subsection{Compared With Self-Implemented MSDF}
\vspace{-10pt}
\begin{figure}[!ht]
  \centering
  \includegraphics[width=1.0\textwidth]{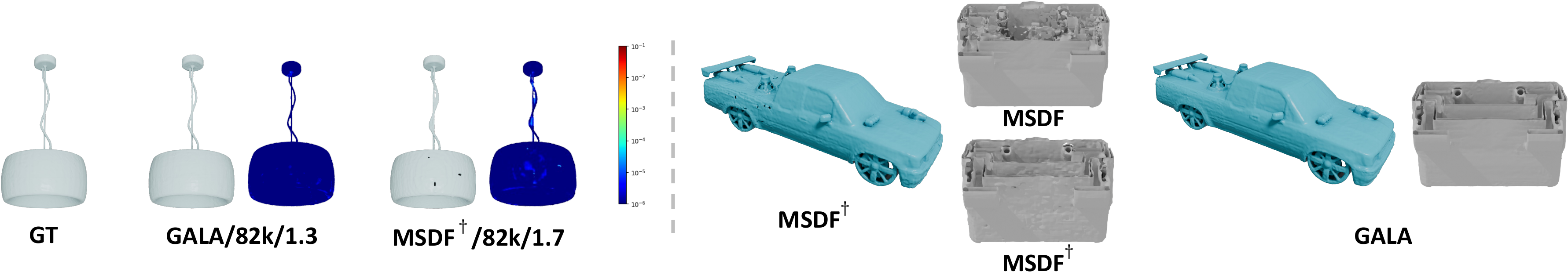} 
  \caption{Compare GALA with our self-implemented Mosaic-SDF (MSDF$^\dagger$).}
  \label{fig:compared_with_self_msdf}
\end{figure}
\vspace{-10pt}

Based on our own GALA implementation, we also implemented our own version of Mosaic-SDF (MSDF), where we tried to strictly follow its original setting, such as $m=7$ for each of the axis-aligned and isotropic grid sampling. Firstly, as we can see from the left part of~\autoref{fig:compared_with_self_msdf}, with the same low parameter budget (82k) and even the additional quantization step, our GALA represents much more faithful geometric structures such as smooth lampshade and accurate strings than MSDF, CD ($\times 10^{-3}$) 1.3 vs. 1.7. Secondly, in the right part of~\autoref{fig:compared_with_self_msdf}, our reimplemented MSDF shows a smoother internal structures than the officially provided one but have more uneven exterior surfaces.

\subsection{Time Measurement of \acro Fitting}
\label{subsec:time_measurement}
\begin{figure}[!ht]
    \centering
    \begin{subfigure}[b]{0.45\textwidth}
        \centering
        \includegraphics[width=\textwidth]{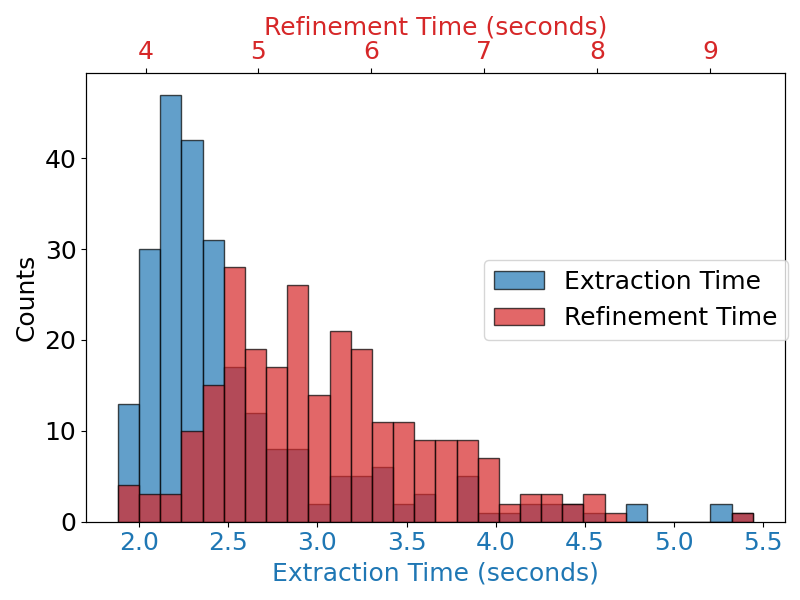}
        \caption{ShapeNet Airplane: \textcolor[rgb]{0.3843, 0.6235, 0.7922}{$\mathbf{2.56}\pm1.30$}, \textcolor[rgb]{0.9568, 0.3647, 0.3921}{$\mathbf{5.60}\pm1.86$}}
    \end{subfigure}
    \hspace{0.05\textwidth}
    \begin{subfigure}[b]{0.45\textwidth}
        \centering
        \includegraphics[width=\textwidth]{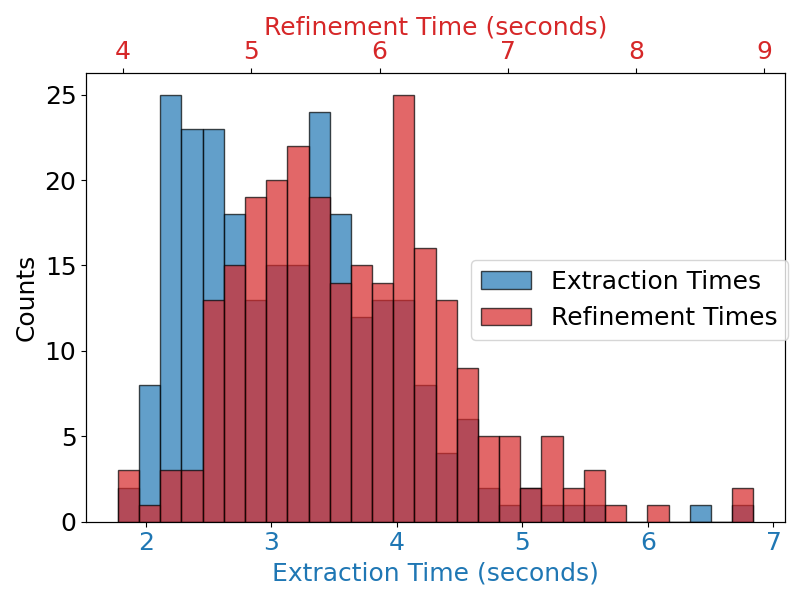}
        \caption{ShapeNet Chair: \textcolor[rgb]{0.3843, 0.6235, 0.7922}{$\mathbf{3.17}\pm1.66$}, \textcolor[rgb]{0.9568, 0.3647, 0.3921}{$\mathbf{5.79}\pm1.67$}}
    \end{subfigure}
    \caption{The fitting time statistics of \acro on ShapeNet Airplane and Chair class.}
    \label{fig:time_measurement}
\end{figure}
\vspace{-10pt}

In this section, we present detailed time measurements of our \acro fitting process. The measurements were conducted under the configuration of 6 virtualized logical cores (hyper-thread) of AMD EPYC 7413 @2.65GHz and 1 Nvidia A100. We show the fitting time statistics plots as in~\autoref{fig:time_measurement} of 250 objects from ShapeNet Airplane and ShapeNet Chair respectively. The extraction (\textcolor[rgb]{0.3843, 0.6235, 0.7922}{blue}) time includes all initialization and local adaptive grid extraction process, namely all the procedures other than the refinement. Based on the comprehensive statistics shown in \autoref{fig:time_measurement}, we show that our implementation indeed achieves very fast fitting in less than 10 seconds overall, as stated previously. Additionally, the peak GPU memory usage during \acro fitting is recorded at less than 800MB.

\subsection{Novelty of Generated Results}

\begin{wrapfigure}[7]{l}{0.4\textwidth} 
\vspace{-18pt}
  \centering
  \includegraphics[width=0.4\textwidth]{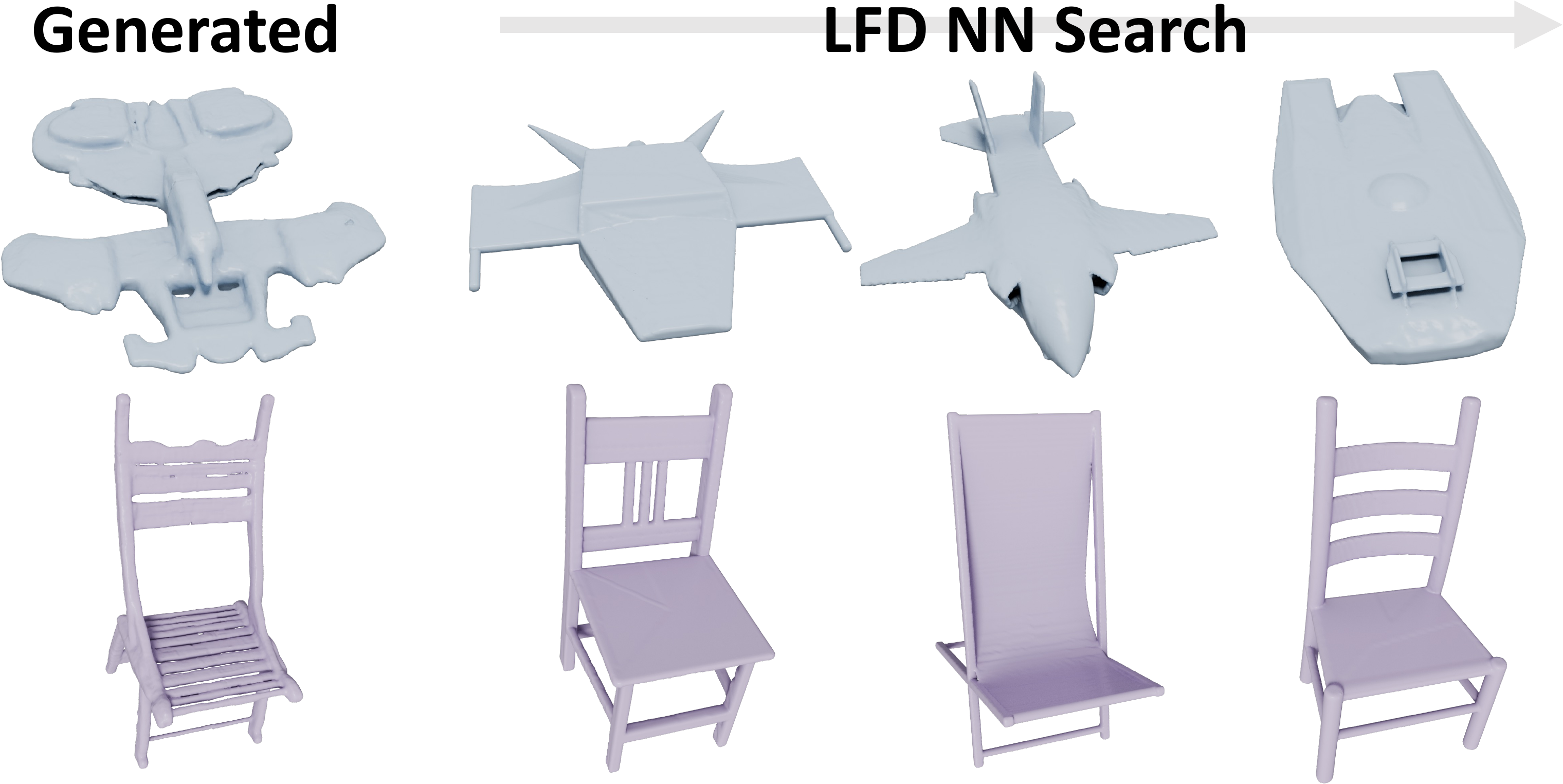} 
  \captionsetup{font=scriptsize}
  \caption{Two generated shapes and three visually closest (left to right) training shapes retrieved using LFD.}
  \label{fig:nn_search_short}
\end{wrapfigure}

In~\autoref{fig:nn_search_short}, 
we show two shapes generated by our method and three closest training shapes retrieved via nearest neighbour search (NN) using LFD~\cite{chen2003visual}. 
We can see that our generations have detailed and distinct geometric structures from those in the training set, demonstrating our method's generative novelty. 

\vspace{10pt}

\subsection{Application: Autocompletion}

\begin{figure}[!ht]
    \centering
    \includegraphics[width=1.0\linewidth]{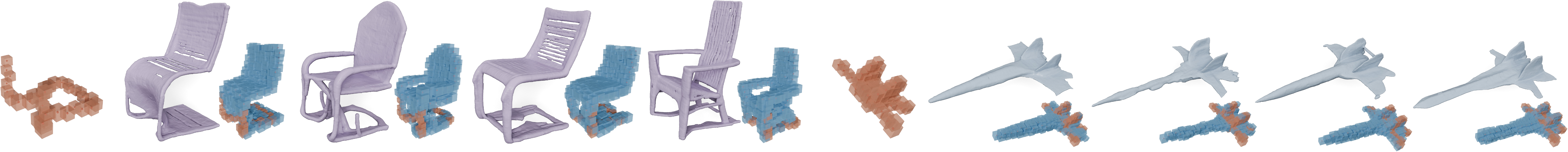}
    \captionsetup{font=small}
    \caption{Given the partial root voxels (\textcolor[rgb]{0.9215686,0.368627,0.203922}{orange}), we autocomplete the rest (\textcolor[rgb]{0.20392157, 0.5764709, 0.9215686}{blue}) and generate geometries.}
    \label{fig:autocomplete}
\end{figure}

Due to the hierarchical structure of \acro and the cascaded generation pipeline, we are able to roughly outline the objects to be generated using coarse and partial root voxels. Based on the work RePaint~\cite{lugmayr2022repaint}, we autocomplete the given root voxels by substituting intermediate results at timesteps with the given ones during denoising process. See~\autoref{fig:autocomplete} for the results.

\vspace{-10pt}

\subsection{Texturing the Generated Mesh}
\begin{wrapfigure}[9]{l}{0.3\textwidth} 
\vspace{-18pt}
  \centering
  \includegraphics[width=0.28\textwidth]{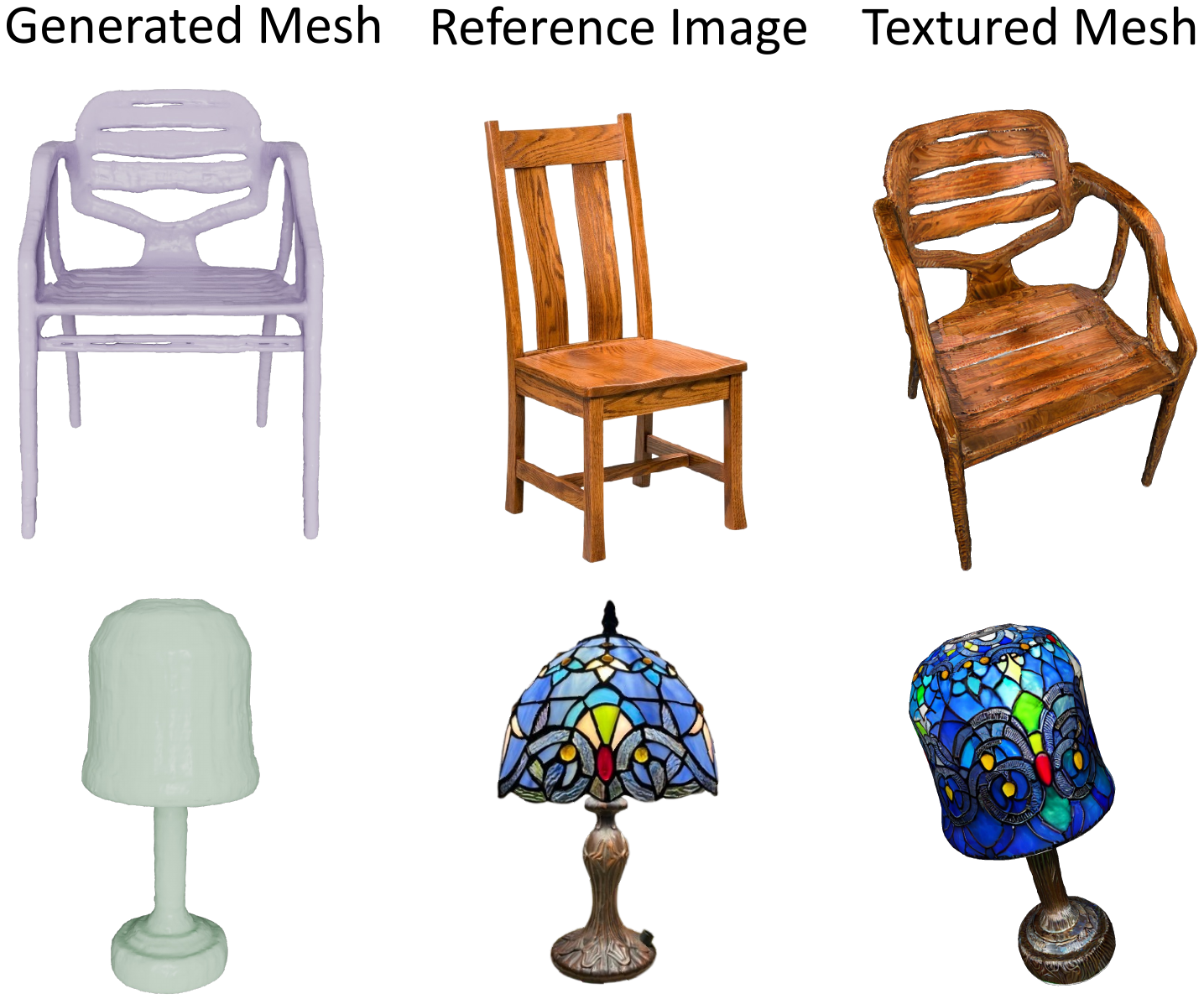} 
  \captionsetup{font=scriptsize}
  \caption{Texturing  generated meshes. }
  \label{fig:texture_show}
\end{wrapfigure}
First carving the mesh and then adding textures is a standard workflow in 3D asset creation within the industry. It allows flexibility of creating various textures while reusing the fine meshes already crafted. Many previous works~\cite{zhang2024clay,perla2024easitex,cao2023texfusion,chen2023text2tex,richardson2023texture} in deep learning also adopt similar strategies. Therefore, to show our generated meshes can be easily textured, in ~\autoref{fig:texture_show}, we use Easi-Tex~\cite{perla2024easitex} to texture examples of our generated meshes by reference images.


\vspace{2pt}

\section{Conclusion and Future Work}
\label{sec:future}

We present GALA, geometry-aware local adaptive grids for detailed 3D generation. We demonstrate that \acro can represent detailed geometries efficiently with fewer parameters. Additionally, we propose a cascaded generation approach for \acro that produces detailed geometry. \dd{We also deliver an efficient C++/CUDA implementation of \acro that completes fitting in seconds}, significantly enhancing the practical usage of \dd{our method}. For future work, compact graph representations~\cite{bouritsas2021partition,jo2022score,jang2023hierarchical} may facilitate tree compression \dd{within} the forest and further ease the generation step. In addition, we show failure cases of imperfect root voxel generation in~\autoref{sec:failure_cases}, which we may improve that using better diffusion training scheme in the near future. Also of interests is to remove the watertight constraint on the input meshes and extend GALA and the ensuring generation to unsigned distance fields (UDFs).

\clearpage
{\small
\bibliographystyle{iclr2025_conference}
\bibliography{main}
}

\appendix
\appendix
\title{Appendix}
\renewcommand{\thesection}{\Alph{section}}
\setcounter{section}{0}


\section{Implementation Details}
\subsection{Flipping Signs of the Interior Parts}
\label{sec:flip_sign}
\begin{wrapfigure}[9]{l}{0.5\textwidth} 
  \centering
  \includegraphics[width=0.48\textwidth]{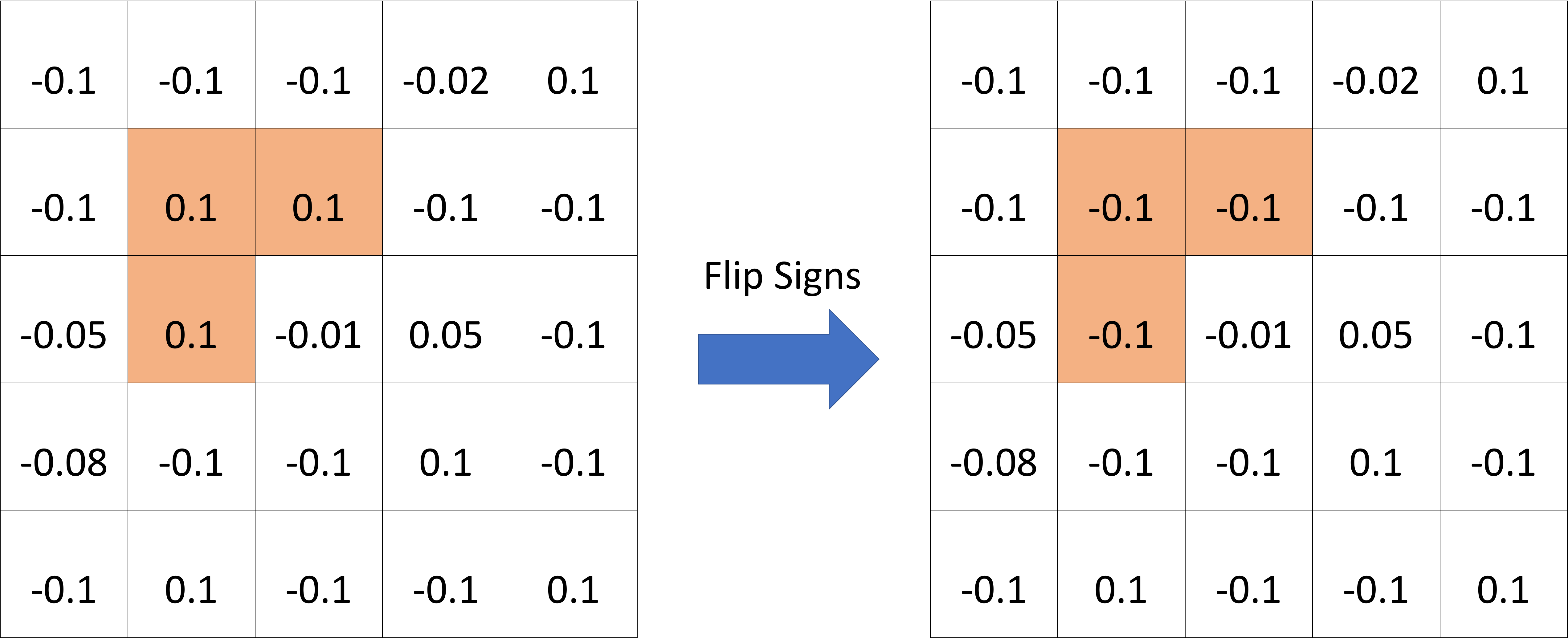} 
  \caption{Flip signs of 0.1 \textit{islands} (connected components) surrounded by negatives.}
  \label{fig:flip_sign}
\end{wrapfigure}
We will flip the signs of \textbf{connected components of 0.1 surrounded by negative values} in a SDF slice $S\in\mathbb{R}^{\ell \times \ell}$ extracted from \acro, where $\ell$ is the extracted SDF grid resolution. It is a Depth-first Search (DFS) algorithm as described in \autoref{alg:flip_sign}. This post-processing procedure will eliminate redundant geometry blobs in interior parts of objects as illustrated in~\autoref{fig:flip_sign}.

\vspace{30pt}

\begin{algorithm}
\caption{Flip Signs of the Interior Parts}
\label{alg:flip_sign}
\begin{algorithmic}[1]
\State \text{Input: An OFALG, $y$ and $\ell$} 
\State Extract SDF slice $S\in\mathbb{R}^{\ell\times \ell}$ at $y$ \Comment{According to Eq (2,3) of the main text.}
\State Initialize visited matrix $V$ of size $\ell \times \ell$ to false
\State Define directions $D = \{(0,1), (0,-1), (1,0), (-1,0)\}$
\For{$i = 0$ to $\ell - 1$}
    \For{$j = 0$ to $\ell - 1$}
        \If{$\lvert S[i, j] - 0.1 \rvert < \text{EPS} \And \neg V[i, j]$} \Comment{Check possible islands of 0.1 SDF values.}
            \State $stack \gets \emptyset$
            \State $stack.\text{push}((i, j))$
            \State $V[i, j] \gets \text{true}$
            \State $isSurrounded \gets \text{true}$
            \State $positions \gets \emptyset$
            \While{not $stack.\text{empty}()$}
                \State $(x, y) \gets stack.\text{pop}()$
                \State $positions.\text{add}((x, y))$
                \ForAll{$(dx, dy) \in D$}
                    \State $(nx, ny) \gets (x + dx, y + dy)$
                    \If{$nx < 0 \lor nx \geq \ell \lor ny < 0 \lor ny \geq \ell$} \Comment{Boundary condition.}
                        \State $isSurrounded \gets \text{false}$
                        \State \textbf{continue}
                    \EndIf
                    \If{$\lvert S[nx, ny] - 0.1 \rvert < \text{EPS} \And \neg V[nx, ny]$} \Comment{Push 0.1 to the island.}
                        \State $stack.\text{push}((nx, ny))$
                        \State $V[nx, ny] \gets \text{true}$
                    \EndIf
                    \If{$S[nx, ny] \geq 0 \And \lvert S[nx, ny] - 0.1 \rvert > \text{EPS}$} \Comment{Check other positives.}
                        \State $isSurrounded \gets \text{false}$
                    \EndIf
                \EndFor
            \EndWhile
            \If{$isSurrounded$}
                \ForAll{$(x, y) \in positions$}
                    \State $S[x, y] \gets -S[x, y]$ \Comment{Flip the sign of the 0.1 island.}
                \EndFor
            \EndIf
        \EndIf
    \EndFor
\EndFor
\end{algorithmic}
\end{algorithm}

\newpage

\subsection{Data Preprocessing}
\label{sec:data_processing}

\begin{figure}[ht]
    \centering
    \includegraphics[width=1.0\textwidth]{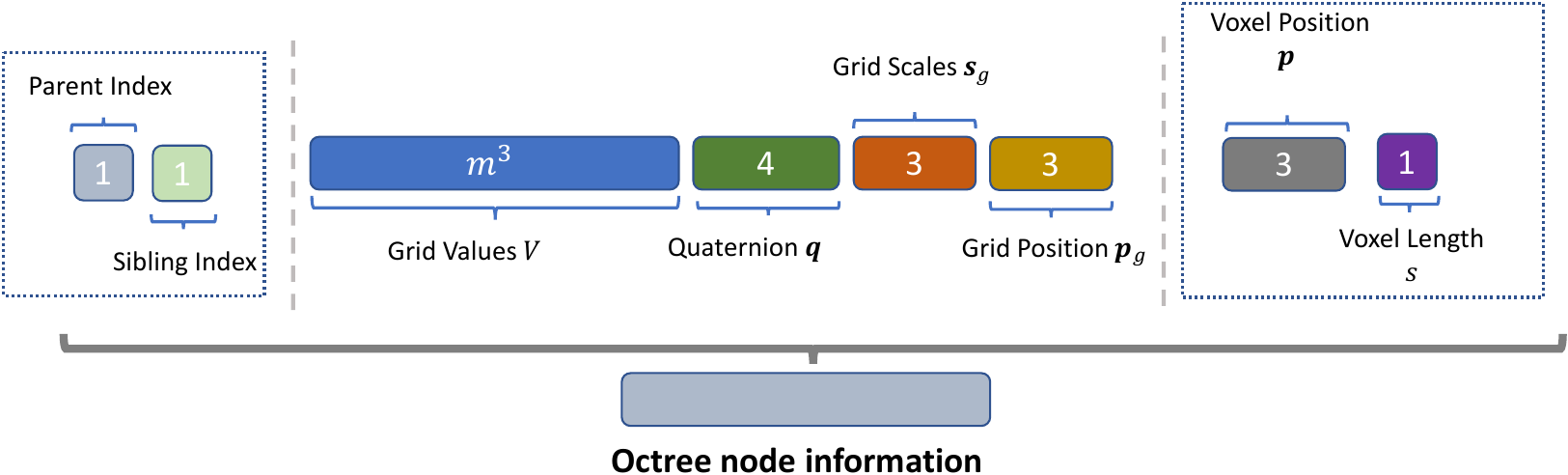}
    \caption{\textbf{Octree Node Information}. \textbf{Left}: the \textit{Parent Index} $\in [0, \dots, N_0\times 8 ^{l-1}-1]$ and \textit{Sibling Index}$\in[0,\dots,7]$, $l$ is the tree level and $N_0$ is the tree root number. When $l=0$, placeholder values are put on the two indices; \textbf{Middle}: local adaptive grid related information, the scales $\mathbf{s}_g$ and position $\mathbf{p}_g$ of the grid are stored; \textbf{Right}: the values of root voxels $\mathbf{p},s$.}
    \label{fig:data_composition}
\end{figure}

For each root or non-empty leaf tree node, information will be saved as illustrated in~\autoref{fig:data_composition}. The voxel position $\mathbf{p}$ and scale $s$ (\autoref{fig:data_composition}.\textbf{Right}) would only be recorded and generated at the root level. The $\mathbf{p},s$ of subsequent levels of tree nodes will be deduced deterministically from the root voxels. The parent index and sibling index (\autoref{fig:data_composition}.\textbf{Left}) is only recorded for recover the tree structure and will not be explicitly generated. Quaternion is used to represent the orientation of the local grid. Moreover, local grid locations and scales are stored and generated. All the quantities within the tree node vector will be normalized to [0,1] by $\min,\max$ statistics calculated within each data domain ($\mathbf{p},s,\mathbf{p}_g$,etc) of specific datasets.

\begin{figure}[ht]
    \centering
    \includegraphics[width=0.8\textwidth]{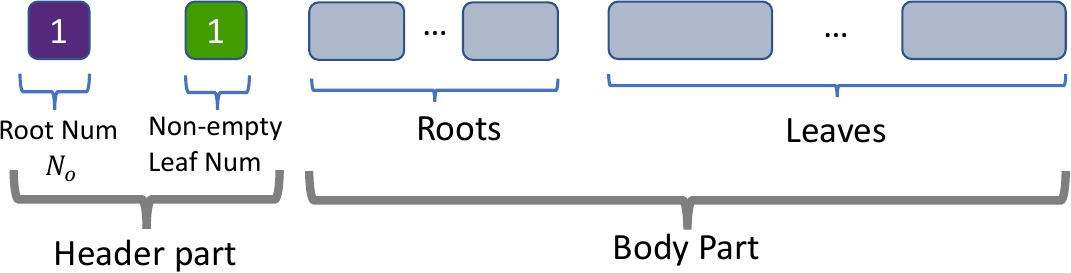}
    \caption{The data format of \acro to be flattened and stored.}
    \label{fig:binary_format}
\end{figure}

The \acro of each object will be flattened and stored as the format illustrated in~\autoref{fig:binary_format}.

\vspace{-20pt}

\subsection{Details of the Pure C++/CUDA Implementation and Code Release}
\label{sec:details_of_the_pure_implementation}
\begin{figure}[!ht]
  \centering
  \includegraphics[width=0.9\textwidth]{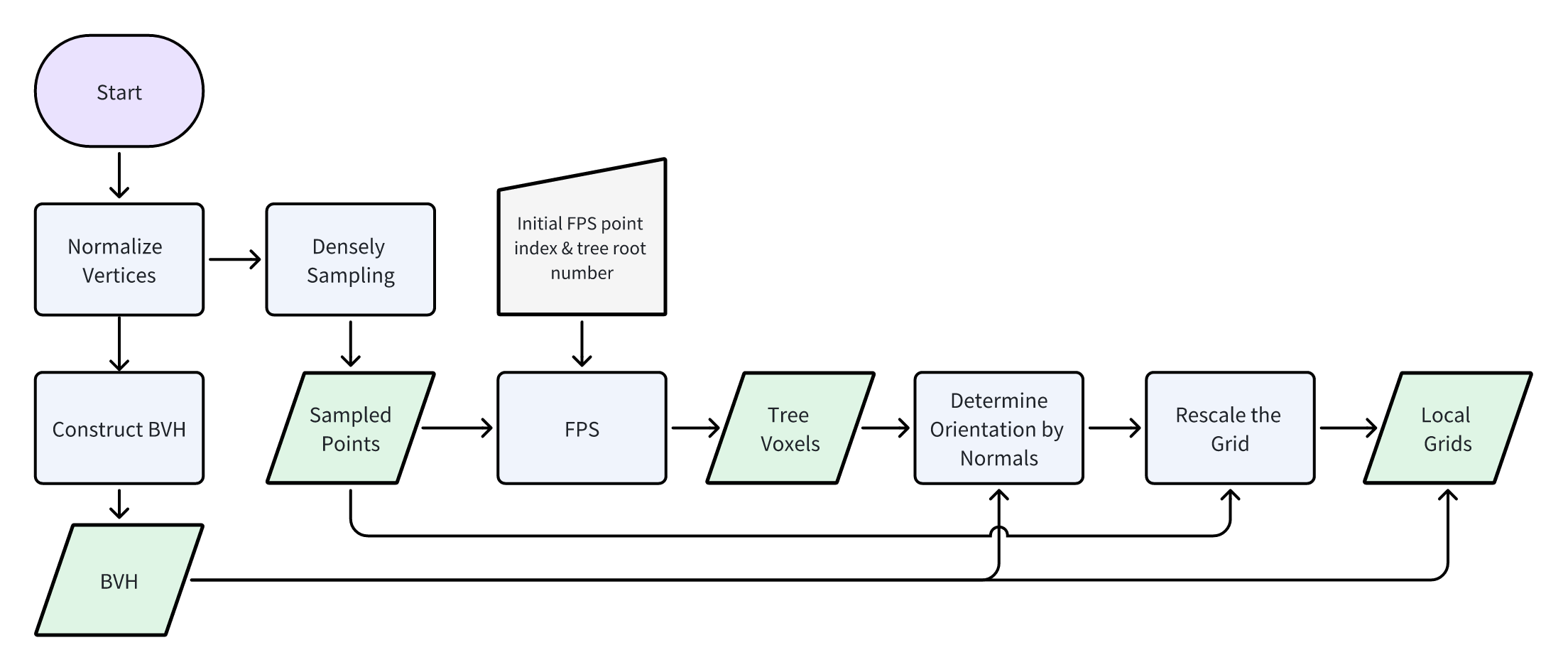} 
  \caption{The algorithm flowchart of our \acro extraction.}
  \label{fig:OFALG_flow}
\end{figure}

We show the algorithm flowchart~\autoref{fig:OFALG_flow} of the initialization stage of \acro, which is built upon CUDA, Libtorch, a third-party R-Tree based SDF query library\footnote{https://github.com/sxyu/sdf} and few head only utility libraries such as C++ argparse\footnote{https://github.com/p-ranav/argparse}. We will dig into the details of the implementation step by step as follows.

\textbf{Normalize Vertices} We shift the vertices by the center location ($(\frac{x_{\text{max}}+x_{\text{min}}}{2},\frac{y_{\text{max}}+y_{\text{min}}}{2},\frac{z_{\text{max}}+z_{\text{min}}}{2})$) and then scale them by the diagonal distance ($\sqrt{(x_{\text{max}} - x_{\text{min}})^2 + (y_{\text{max}} - y_{\text{min}})^2 + (z_{\text{max}} - z_{\text{min}})^2}$).

\textbf{Construct Bounding Volume Hierarchy (BVH)} In order to quickly search relevant mesh information in the space, we construct Bounding Volume Hierarchy (BVH) on GPU parallelly via radix tree according to~\cite{karras2012maximizing}. We encountered some bug when doing the GPU parallel bottom-up AABB (Axis-Aligned Bounding Box) construction of the radix tree where very few random items of AABB ($\sim$ 0.01\%) at z-dimension were randomly corrupted for each run. To circumvent this, we copy the whole tree back to CPU, conduct a DFS post-order traverse to construct the AABB and copy the results back to GPU.

\textbf{Densely Sampling Points On Meshes} We sample $10 \times N_t$ points on meshes where $N_t$ is the number of triangles of the input mesh. We randomly sample points on triangles using barycentric coordinates. The number of points sampled at each triangle is proportional to its surface area.

\textbf{Initial Point of Farthest Point Sampling} As described in the main text, in order to augment our representation, the extraction algorithm also accepts the index of the initial FPS point. This enables possible data augmentation of \acro representation by choosing different initial FPS points for the same input mesh.

\textbf{Determining Orientation by Normals} As detailed in the main text, we estimate the orientation of the local grid using the eigenvectors of the covariance matrix of the bounded normal vectors. We get the eigenvectors in each local grid in parallel on each GPU thread using Jacobi eigenvalue algorithm~\cite{golub2013matrix}, where the maximum iteration is set as 50 and the tolerance for maximum off-diagonal value is $1e^{-10}$. 

\subsection{More details on Cascaded Generation}
\label{sec:more_details_on_generation}
\begin{figure}[!ht]
    \centering
    \includegraphics[width=1.0\linewidth]{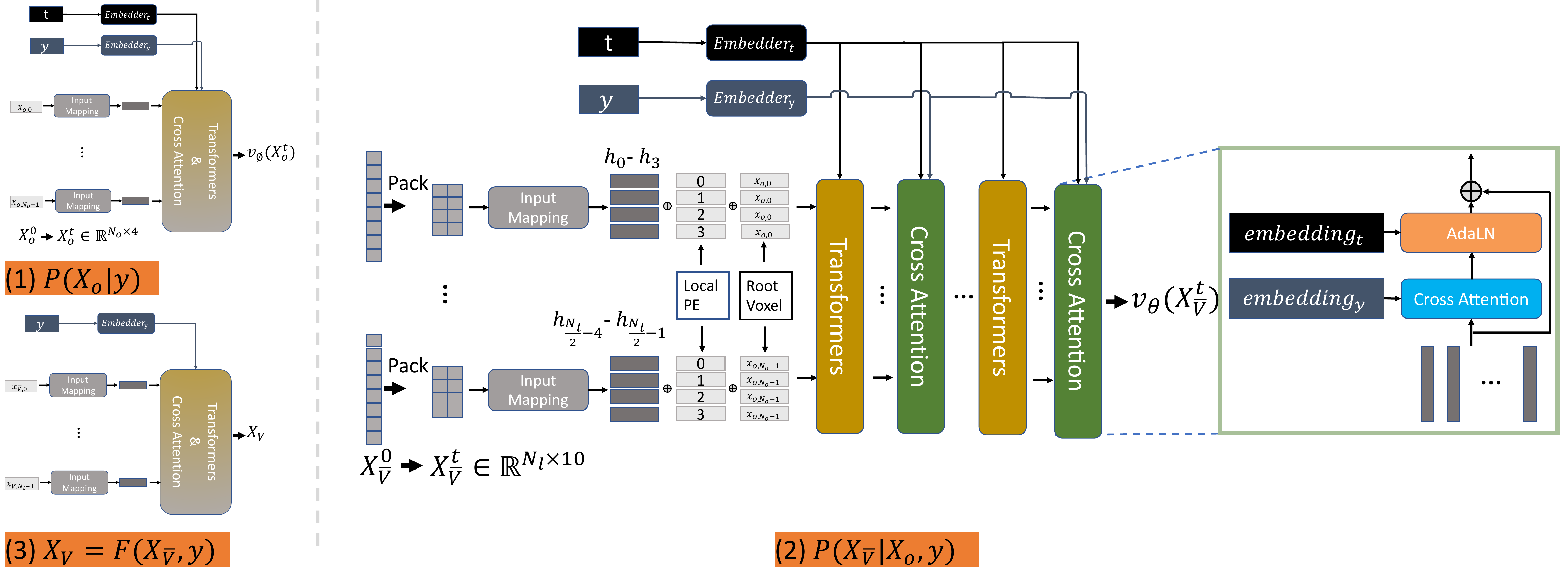}
    \caption{The network details of the cascaded generation pipeline, composed by steps: (1) root voxel diffusion $P(X_o|y)$ by diffusion; (2) Local adaptive grid configuration generation $P(X_{\bar{V}}|X_o,y)$ by diffusion; (3) Local adaptive grid value prediction $X_V=F(X_{\bar{V}},y)$ trained by regression.}
    
    \label{fig:generation_network_details}
\end{figure}

As shown in~\autoref{fig:generation_network_details}, our generation networks are based on transformer networks~\cite{vaswani2017attention}. The root voxel conditions and local positional encodings of the tree are injected in an in-context manner through addition, while the timestep condition is injected using the adaLN-Zero block from DiT~\cite{peebles2023scalable}, and the object class condition is injected via cross-attention. Additionally, noise augmentation is applied, as in previous works~\cite{ho2022cascaded,xu2024brepgen,saharia2022photorealistic}, to mitigate the impact of the training/inference discrepancy in cascaded generation. Furthermore, zero-terminal~\cite{lin2024common} is adopted to ensure the last timestep of diffusion starts from a zero signal-to-noise ratio (SNR). Classifier-free guidance~\cite{ho2022classifier} is used, with an additional null label trained as $y$ during diffusion training, and $w=0$ is applied during classifier-free guidance in diffusion inference.

\section{Other Ablation Studies}
%

\subsection{More Illustration on w/o $\mathcal{A}$}
\label{sec:more_on_wo_A}

\begin{figure}[ht]
\centering
\begin{subfigure}{\textwidth}
  \centering
  \includegraphics[width=0.95\linewidth]{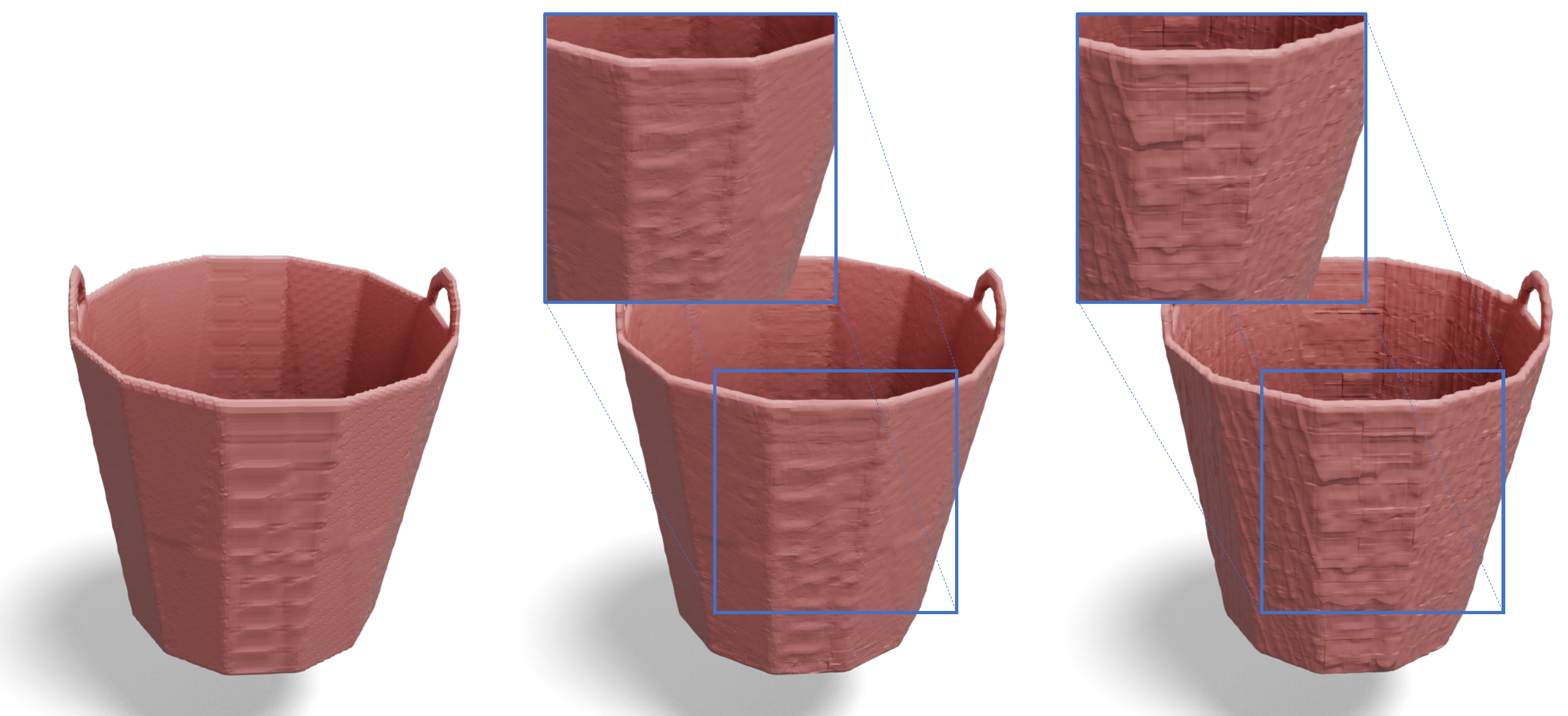}
  \label{fig:sub1}
\end{subfigure}%
\vspace{0cm} 

\begin{subfigure}{\textwidth}
  \centering
  \includegraphics[width=0.95\linewidth]{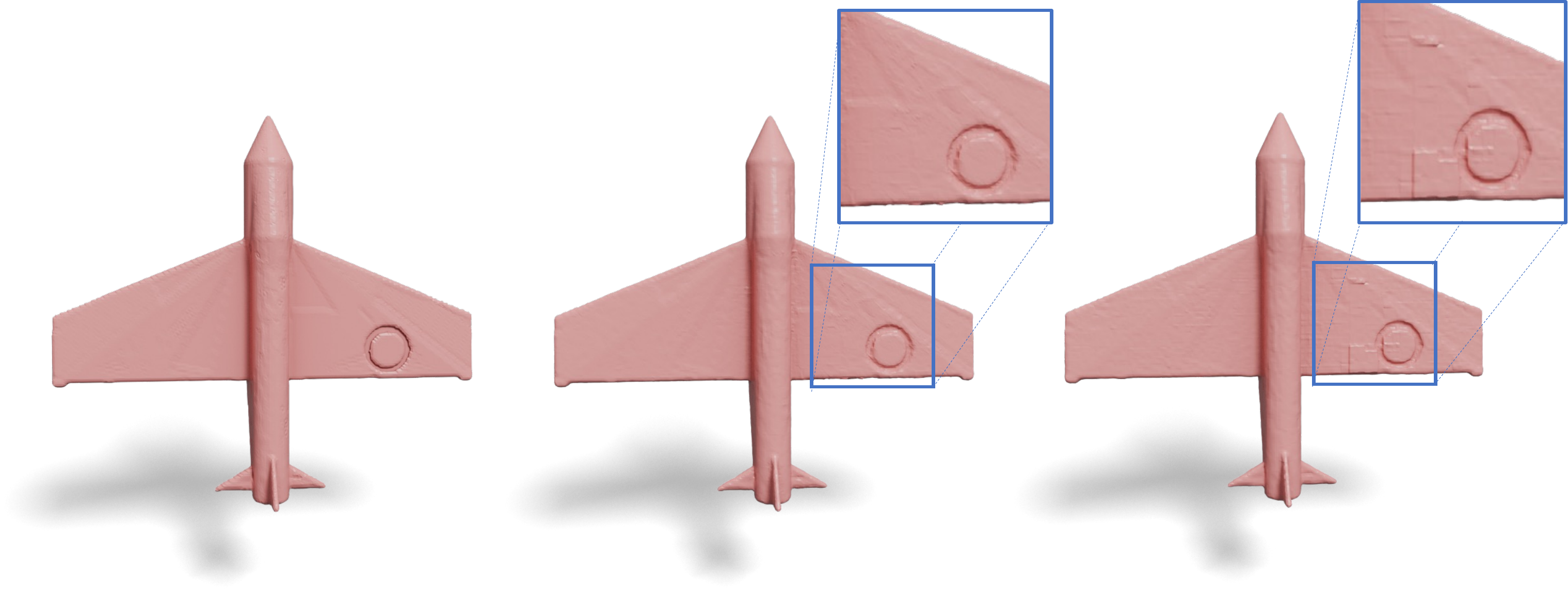}
  \label{fig:sub2}
\end{subfigure}
\vspace{-0.5cm} 

\begin{subfigure}{\textwidth}
  \centering
  \includegraphics[width=0.95\linewidth]{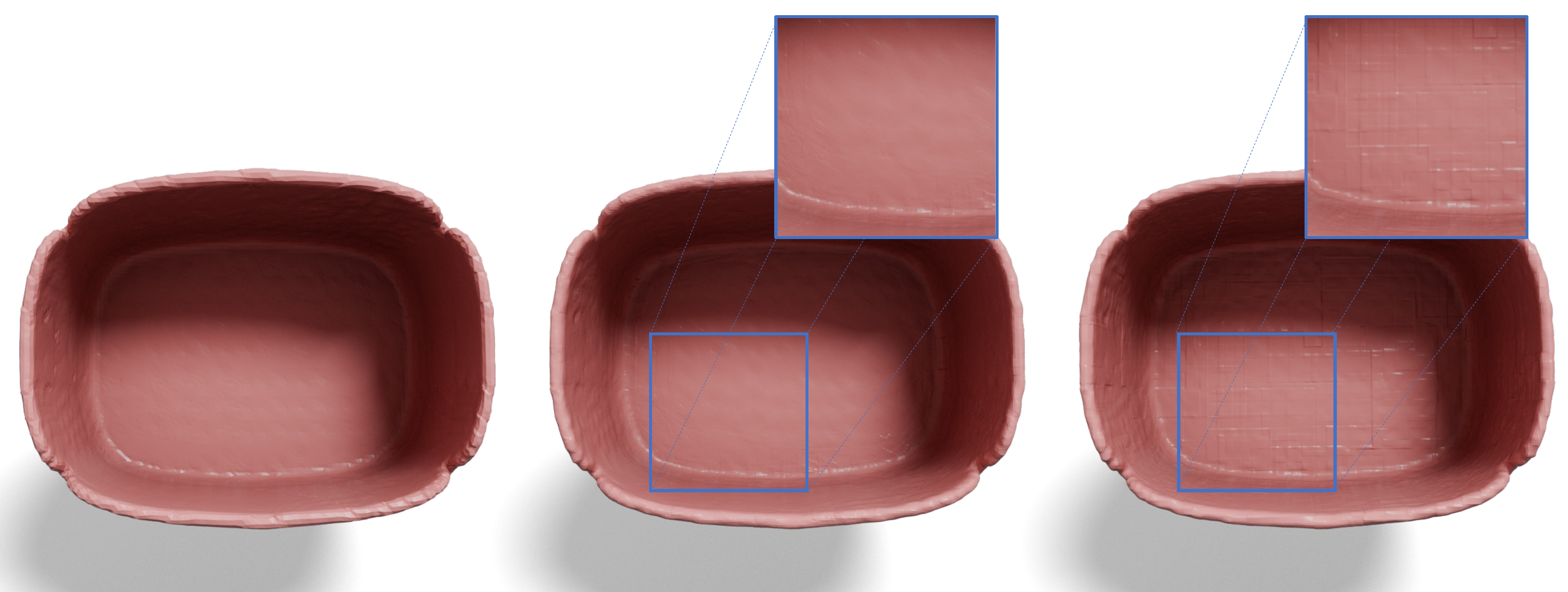}
  \label{fig:sub3}
\end{subfigure}

\caption{Qualitative comparison between \acro (ours, \textbf{Middle}) and \acro without anisotropic local grids (w/o $\mathcal{A}$, \textbf{Right}). The $\textbf{Left}$ is the ground truth (GT).}
\label{fig:woA_qualitative_comparison}
\end{figure}

As we can see from~\autoref{fig:woA_qualitative_comparison}, since the anisotropic local grid setting ($\mathcal{A}$) enables close and adaptive sampling on object surfaces, the full setting of \acro captures much more accurate and artifact-free surfaces than w/o $\mathcal{A}$.

\newpage

\subsection{Mesh Extraction Efficiency}
As in the Mosaic-SDF~\cite{yariv2023mosaic} paper, we report the mesh extraction time at resolution $256^3$ and $512^3$ on NVIDIA A100 in~\autoref{tab:extraction_efficiency}. We show that with our implementation, the mesh extraction process of ours is very efficient. To further dissect the running time of our mesh extraction process, we show the flip-sign algorithm on CPU($^\dagger$) consists half of the time. Further improvement can be done by adopting GPU graph traversal algorithm like~\cite{merrill2012scalable}. The configuration is the same as stated in~\autoref{subsec:time_measurement}.

\begin{table}[ht]
\centering
\resizebox{\textwidth}{!}{
\begin{tabular}{@{}lcclcc} 
\thickhline
\addlinespace[4pt] 
Method & Resolution &  Time(s) & Method & Resolution & Time(s) \\
\midrule
3DIGL~\cite{zhang20223dilg} & $256^3$ & 18.44 & 3DIGL~\cite{zhang20223dilg} & $512^3$ & 159.56 \\
S2VS~\cite{zhang20233dshape2vecset} & $256^3$ & 5.62$^\ast$ &
S2VS~\cite{zhang20233dshape2vecset} & $512^3$ & 24.49$^\ast$
\\
Mosaic-SDF~\cite{yariv2023mosaic} & $256^3$ & 2.74 & Mosaic-SDF~\cite{yariv2023mosaic} & $512^3$ & 21.84 \\
Ours    & $256^3$ & \makecell{\textbf{2.39} \\
                              \scriptsize (1.26 + 1.23$^\dagger$)} & Ours & $512^3$ &
                              \makecell{\textbf{18.73}\\
                              \scriptsize (8.79 + 9.94$^\dagger$)} \\                        
                              
\addlinespace[4pt] 
\thickhline
\end{tabular}
}
\vspace{5pt}
\caption{Extraction efficiency comparison with Mosaic-SDF and other baselines on NVIDIA A100. $^\dagger$ means the running time of the DFS-based flip-sign algorithm detailed in \autoref{sec:flip_sign} on CPU. $^\ast$ means when measuring the inference time of S2VS, we conduct queries of $256^2$ or $512^2$ slice every time and loop for the whole grid in order to avoid OOM issue on GPU.}
\label{tab:extraction_efficiency}
\end{table}

\subsection{Storage Efficiency}
\label{sec:storage_efficiency}
\begin{wrapfigure}[15]{l}{0.5\textwidth} 
  \centering
  \includegraphics[width=0.48\textwidth]{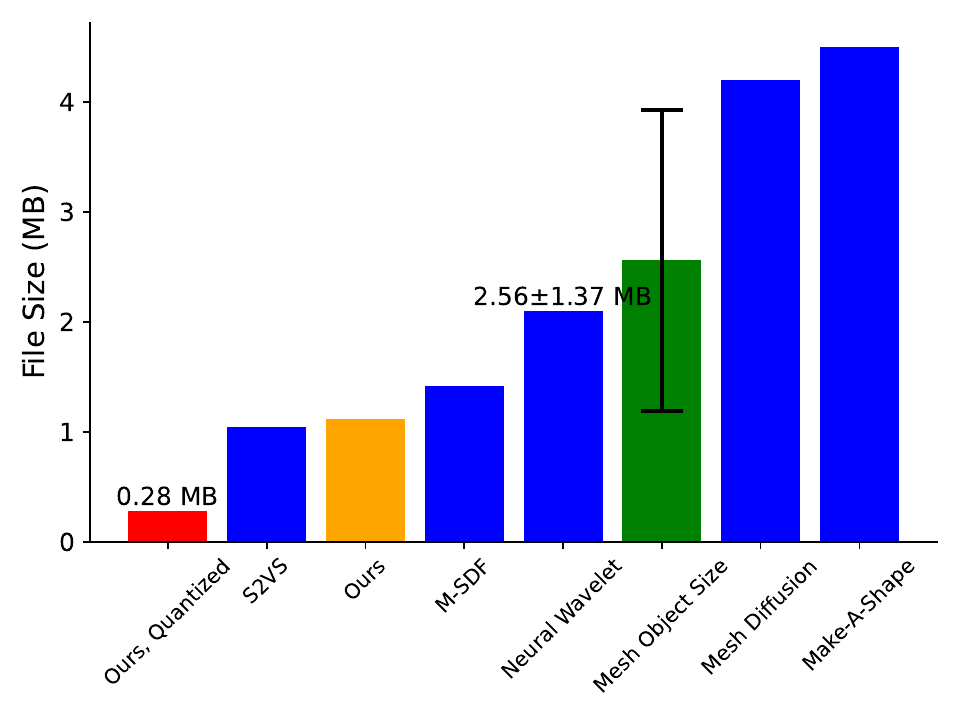} 
  \caption{Comparison of file sizes of each representation.}
  \label{fig:file_size}
\end{wrapfigure}
In \autoref{fig:file_size}, we compare the file sizes of different representations in order to illusrate the high storage efficiency of our \acro. The representations to be compared in this figure, from the left to the right, are ours (\acro), S2VS~\cite{zhang20233dshape2vecset}, M-SDF~\cite{yariv2023mosaic}, Neural Wavelet~\cite{hui2022neural}, raw object mesh files, Mesh Diffusion (DMTet)~\cite{liu2023meshdiffusion}, Make-A-Shape (Packed wavelet coefficients)~\cite{hui2024make}. Raw object mesh sizes are measured on the converted water-tight meshes on ShapeNet Airplane. Each file of ours occupies only 0.28MB disk space, which is a practical benefit when processing large scale datasets.

%
%

\vspace{15pt}

\section{Failure Cases}
\label{sec:failure_cases}
\begin{wrapfigure}[12]{l}{0.3\textwidth} 
  \centering
  \includegraphics[width=0.3\textwidth]{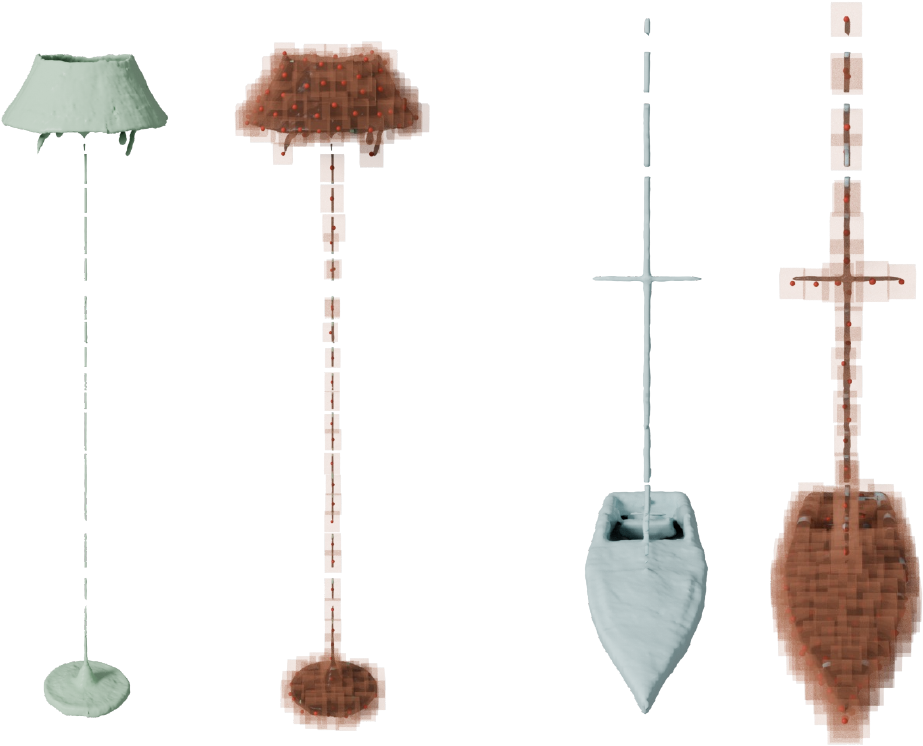} 
  \caption{Examples of typical failure cases. Root voxels are visualized on the right.}
  \label{fig:failure_case_show}
\end{wrapfigure}
At this section, we show examples of typical failure cases. As we can see from~\autoref{fig:failure_case_show}, typical failure cases root in imperfect root voxel generation, which is the first step of our cascaded generation pipeline. Failed generation of properly \textit{connected} root voxels would result in unconnected geometries. It will be beneficial to study how to enhance the root voxel generation, for example using more advanced diffusion training schemes. On the other hand, it is worth to note, although some of the root voxels are not properly connected, the final generated geometries are very well aligned, fine and reasonable, implying the power of \acro and our cascaded generation pipeline.

\end{document}